\PassOptionsToPackage{table}{xcolor} %
\documentclass{article} %
\usepackage{iclr2027_conference,times}

\usepackage{amsmath,amsfonts,bm}

\def\eqref#1{equation~\ref{#1}}
\def\Eqref#1{Equation~\ref{#1}}

\def\1{\bm{1}}

\DeclareMathAlphabet{\mathsfit}{\encodingdefault}{\sfdefault}{m}{sl}
\SetMathAlphabet{\mathsfit}{bold}{\encodingdefault}{\sfdefault}{bx}{n}

\newcommand{\R}{\mathbb{R}}

\newcommand{\KL}{D_{\mathrm{KL}}}

\DeclareMathOperator*{\argmax}{arg\,max}

\usepackage{float} %
\usepackage{hyperref}
\usepackage{url}
\usepackage{graphicx}
\usepackage{wrapfig}
\usepackage{algorithm}
\usepackage{algpseudocode}
\usepackage{booktabs}
\usepackage{array}
\usepackage{xcolor}
\usepackage{colortbl} %
\usepackage[normalem]{ulem}
\usepackage{enumitem}
\usepackage{xspace}
\usepackage{cancel}

\usepackage{framed}
\usepackage{makecell}
\usepackage{multirow}
\usepackage{amsthm}
\usepackage{tikz}
\usetikzlibrary{arrows.meta,positioning,decorations.pathmorphing,calc}
\usepackage{pgfplots}\pgfplotsset{compat=1.18}
\definecolor{palLightBlue}{HTML}{9FC7E2}
\definecolor{palBlue}{HTML}{50759F}
\definecolor{palLightOrange}{HTML}{F8B886}
\definecolor{palOrange}{HTML}{E88532}
\definecolor{palLightPurple}{HTML}{D1A5C7}
\definecolor{palPurple}{HTML}{AB779D}
\definecolor{palLightGreen}{HTML}{8EC579}
\definecolor{palGreen}{HTML}{5B984B}

\definecolor{suiteshade}{gray}{0.92}
\definecolor{shadecolor}{gray}{0.95} %
\theoremstyle{plain}
\newtheorem{proposition}{Proposition}

\theoremstyle{definition}
\newtheorem{definition}{Definition}
\newtheorem*{proposition*}{Proposition} %
\theoremstyle{remark}\theoremstyle{plain}
\newcommand{\ours}{D\&R\xspace}

\title{Divide-and-Remember: Recursive Action-Relevant Memory for Long-Horizon VLA Policies}

\author{
Xuehui Yu$^{1}$ \quad Eason Yu$^{2}$ \quad Meiyi Wang$^{3}$ \quad Haozhe Du$^{1}$ \quad Stefano V. Albrecht$^{2}$ \quad Harold Soh$^{1}$ \\[2pt]
$^{1}$National University of Singapore \quad $^{2}$Nanyang Technological University, Singapore \quad $^{3}$LISTENAI \\[2pt]
{\small\texttt{\{yuxuehui, harold\}@nus.edu.sg} \quad \texttt{haozhe\_du@u.nus.edu} \quad \texttt{mywang2@listenai.com}} \\
{\small\texttt{yuyu0027@e.ntu.edu.sg} \quad \texttt{stefano.albrecht@ntu.edu.sg}}
}

\iclrfinalcopy %
\begin{document}
\raggedbottom %
\addtocontents{toc}{\protect\setcounter{tocdepth}{-2}} %

\maketitle
\lhead{Preprint}

\begin{abstract}
Vision--language--action (VLA) models struggle on history-dependent manipulation tasks, where the current observation alone does not determine the action, and the policy needs a memory of the history.  
Existing memory methods decide what to remember by design, for example, keeping frames with large pixel changes, and show inconsistent gains across tasks.
We view what to remember as an optimisation problem. From the POMDP formulation of imitation learning, we show that the optimal memory maximises the conditional mutual information $I(a_t;\,m_t\mid o_t)$ between the action and the memory given the current observation.
Intuitively, this means preserving the action-relevant information in the history that is not already contained in the current observation. Based on our analysis, we propose Divide-and-Remember (\ours), a recursive memory method that learns a memory function $m_t=\mathcal M(h_t)$ and scales to long contexts while staying compute-light.
It involves two strategies: (1) the selection over the full history is divided recursively into subproblems of top-$K$ selection over $2K$ tokens, so that fixed-size, lightweight selectors learned end-to-end support an unbounded history;
(2) all recursion blocks share one selector, which captures the selection rule common to every block and keeps the method efficient.
On RoboMME, a benchmark of $16$ long-horizon manipulation tasks that require remembering when, where, what, and how to act, \ours achieves a state-of-the-art average success rate with consistent gains across all four suites under a budget of only $64$ tokens; real-robot experiments show the same gain.
Code, checkpoints and more results are at \href{https://dnr-memory.github.io/}{dnr-memory.github.io}.
\end{abstract}

\section{Introduction}
\begin{figure}[t]
\centering
\includegraphics[width=\linewidth]{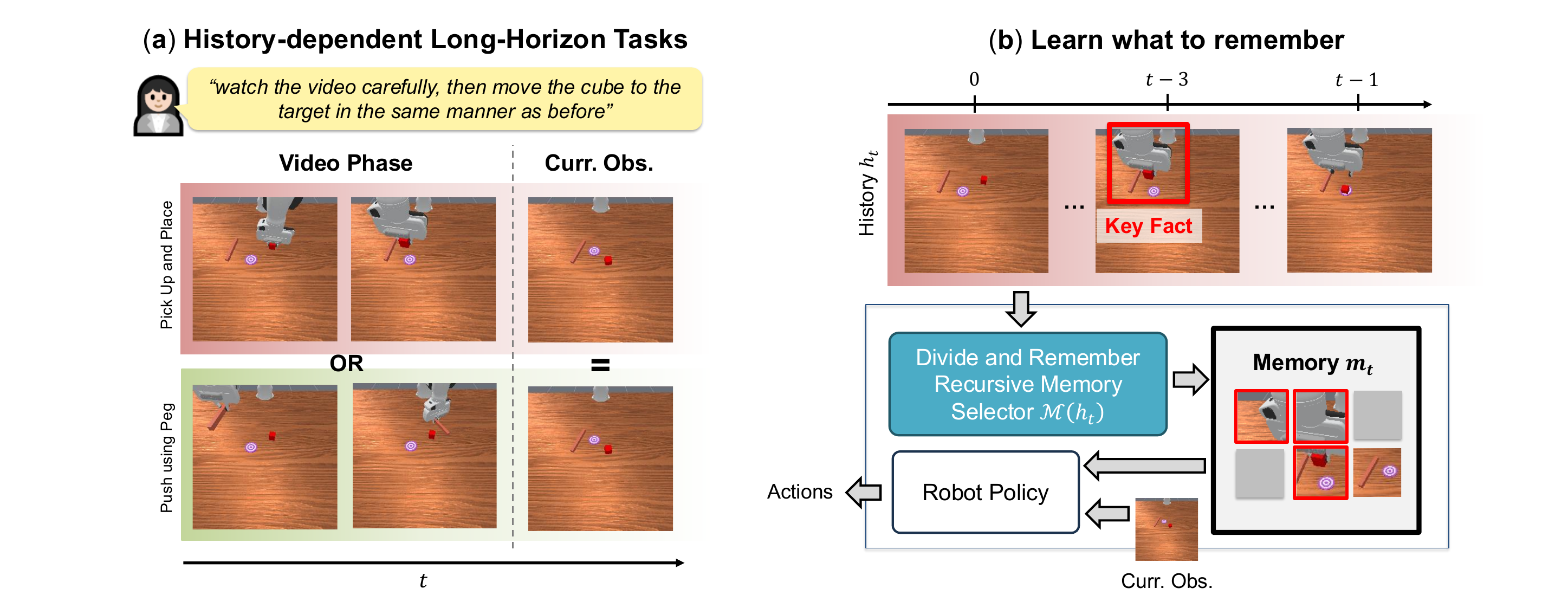}
\caption{\textbf{Memory is a question of what to remember.}
(a) A history-dependent task: ``watch the video carefully, then move the cube to the target in the same manner as before''. Two videos show different manners (pick-and-place versus peg push), yet at time step $t$ the robot sees the \emph{same} observation. A memoryless policy cannot recall how the cube was moved.
(b) We view memory as an optimisation problem: find the memory $m_t=\mathcal M(h_t)$, selected from the full history, that carries the key fact the current observation lacks (here, the grasp pose at $t-3$), so that the policy knows how to move the cube. 
}
\label{fig:intro}
\end{figure}

Vision--language--action (VLA) policies are visuomotor policies trained by imitation learning, a.k.a.\ learning from demonstrations (LfD): a policy is learned from a dataset $\mathcal D$ of demonstrations, each an observation--action sequence $\{(o_t,a_t)\}_{t=1}^{T}$. Powered by large-scale robotic datasets and pretrained vision--language models, VLAs have achieved remarkable progress in robotic manipulation.
However, many long-horizon tasks are partially observable, where the current observation $o_t$ alone cannot determine the correct action. In Figure~\ref{fig:intro}(a), the robot must move a cube ``in the same manner as before'', as shown in a video it has just watched; after either video it faces the \emph{same} observation, and only the history tells it how to act.
When the task's memory length, the shortest history the task requires, exceeds the policy's context length (Definition~\ref{def:long-horizon}), such ambiguity can only be resolved by the history $h_t=\{(o_k,a_k)\}_{k<t}$. In this case, the policy needs \textbf{memory}, the ability to retain and reuse relevant information across time; we refer to this as the \textbf{memory problem}.

A growing body of research has integrated memory into VLAs, differing in what the memory encodes, from latent memory~\citep{koo2026hamlet,memoryvla2025,cronusvla2025} to perceptual memory~\citep{yang2026eventvla,dai2026robomme}, and alleviates the memory problem to some extent. However, what information the memory should carry remains fragmented. 
Latent memory learns a dense compression of the history $h_t$, which tracks temporal facts well (e.g., a count of events) but aligns poorly with the pretrained token space of the VLA and loses spatial detail. 
Perceptual memory keeps a subset of the history chosen by an inductive bias (e.g., the frames with the largest pixel change, or those picked by a VLM verifier); its verbatim snapshots preserve spatial facts (object locations and spatial relations) and object identity, but often miss some key robot behaviour, and the selection is not optimised for control. 
This diversity reflects that existing work designs memory from task-specific~\citep{yang2026eventvla,kemo2026} perspectives, yielding inconsistent gains and a limited understanding of what the optimal formulation of memory is.

Starting from information theory and POMDPs, we cast the memory problem of a visuomotor policy as an optimisation problem: \textbf{\boldmath the optimal memory maximises the conditional mutual information $I(a_t;\,m_t\mid o_t)$ between the action $a_t$ and the memory $m_t$ given the current observation $o_t$} (formal definition in Section~\ref{sec:analysis}); intuitively, it keeps the key facts that $o_t$ lacks but $a_t$ depends on, such as the grasp pose in Figure~\ref{fig:intro}(b). 
This objective gives a unified view of existing memories. 
Perceptual heuristics keep the frames that differ most in pixels. This is a proxy for the conditional entropy $H(o_k\mid o_t)$, i.e., how much a candidate frame $o_k$ differs from the current observation $o_t$. It is only a loose upper bound on what the frame adds, $I(a_t;\,o_k\mid o_t)\le H(o_k\mid o_t)$, and spends the memory budget on information that does not help control, such as when the observation changes for reasons unrelated to the action (e.g., a lighting change).
Latent memory does optimise $I(a_t;\,m_t\mid o_t)$ through its variational lower bound; the action log-likelihood $\mathbb E_{\mathcal D}[\log\pi_\theta(a_t\mid o_t,m_t)]$ is maximised jointly over the policy weights $\theta$ and the memory $m_t$, but it learns $m_t$ in a new latent space and loses the spatial detail that the pretrained tokens of the VLA carry. 
Thus, we ask: {\it Can a memory be trained to maximise $I(a_t;\,m_t\mid o_t)$ directly, without heuristics, while staying in the pretrained token space of the VLA, so that it is optimised for control and still keeps the spatial details?}

We answer this question with \textbf{Divide-and-Remember (\ours)}, which optimises the memory against $I(a_t;\,m_t\mid o_t)$ in the pretrained token space of the VLA, so that choosing $m_t$ becomes a discrete optimisation over the history tokens $\mathcal H_t$. \ours is a recursive memory method, in which the full history is divided recursively into slices of $2K$ tokens and a fixed-size, lightweight selector keeps $K$ tokens of each slice. The $K$ tokens kept from two neighbouring slices are merged into a new $2K$-token slice and selected again (see Figure~\ref{fig:tree}), and the composition of these selections solves the memory selection over the full history. This design lets the history scale up while the learnable parameters stay light. The selector is trained end-to-end by the policy's action loss, the variational lower bound of $I(a_t;\,m_t\mid o_t)$.

We evaluate on RoboMME~\citep{dai2026robomme}, a benchmark of $16$ long-horizon manipulation tasks that each require a different fact from the past. Under a memory budget of only $64$ tokens, \ours reaches a state-of-the-art average success rate of $38.6\%$, against $32.2\%$ for the best latent memory and $17.9\%$ for the memoryless policy. \ours leads on every memory demand of the benchmark, whether the policy has to accumulate a count over repeated events, track an object after it is hidden, refer back to an earlier moment of the episode, or reproduce a demonstrated sequence.
We also evaluate \ours on a real robot over several long-run tasks, with a person intervening mid-task and under perception noise, where \ours achieved the highest success rate.

\section{Related work}

\textbf{Memory-augmented VLAs.} Existing memory methods for VLAs differ in what the memory encodes, and each fixes what to remember by design.
(1) \emph{Latent memory} compresses the history into a compact representation, by merging each step into learnable moment tokens (HAMLET~\citep{koo2026hamlet}) or by a recurrent belief state trained with world-model objectives (RB-VLA~\citep{bagaria2026rbvla}); it tracks temporal facts well but loses spatial detail.
(2) \emph{Perceptual memory} keeps a subset of the raw history chosen by a fixed selection rule: frames a VLM labels as events (EventVLA~\citep{yang2026eventvla}), the patches with the largest pixel change (TokenDrop~\citep{dai2026robomme}), or uniformly sampled frames (FrameSamp~\citep{dai2026robomme}); the snapshots preserve spatial and object detail, but the rule is hand-designed and not optimised for control.
(3) \emph{Parametric memory} folds the history into fast weights by test-time training (TTT)~\citep{sun2024ttt,dai2026robomme}; a fact stored in weights is written by gradient descent, which is slow, and is hard to read back exactly.

\textbf{Routes to the optimal POMDP policy.} Classical solutions differ in which statistic of the history the policy conditions on.
(1) The full history, $\pi(\cdot\mid o_t,h_t)$, by solving the associated dynamic decision network (Figure~\ref{fig:gm}(a)); the policy grows exponentially with the history's length, so this works only for short histories~\citep{poole2023artificial}.
(2) The belief state, the posterior over the current state and a sufficient statistic for prediction, as in RB-VLA~\citep{bagaria2026rbvla}; its recurrent update makes compute and activation memory grow linearly with the episode length and cannot be parallelised over time, which is costly when every step runs a full forward pass of the VLA.

Our analysis gives all of these designs a common objective, $I(a_t;\,m_t\mid o_t)$; \ours is an approximation that optimises it in the pretrained token space, conditioning the policy on a bounded statistic, so that the memory stays small and the facts it keeps can be read back exactly.

\section{Preliminaries} \label{sec:pre}
\textbf{POMDPs and visuomotor policies.} Instead of the state being observable, an agent gets only partial and/or noisy observations of the state before it has to act. Such a dynamic system is defined by a POMDP $\langle\mathcal S,\mathcal A,\mathcal O,\mathcal T,\Omega\rangle$ with states $s_t\in\mathcal S$, actions $a_t\in\mathcal A$, and observations $o_t\in\mathcal O$, as depicted by the decision diagram in Figure~\ref{fig:gm}(a). 
The initial state is drawn from $p(s_0)$, the transition function
$\mathcal T(s_{t+1}\mid s_t,a_t)=p(s_{t+1}\mid s_t,a_t)$ specifies the dynamics,
and the observation function $\Omega(o_t\mid s_t)=p(o_t\mid s_t)$ specifies how observations arise from the state; all three are unknown.

A memoryless visuomotor policy $\pi_\theta(a_t\mid o_t)$, parameterised by $\theta$, allows an agent to choose actions $a_t$ in response to observations $o_t$ to control a dynamic system, such as a robot. In LfD, we do not require any explicit reward function; we assume instead that the demonstrations come from an expert policy \citep{levine2016visuomotor}.
The dynamics and the expert policy together induce a distribution over trajectories, each consisting of $T$ observation--action pairs over a finite horizon, $\tau=\{(o_t,a_t)\}_{t=1}^{T}$. A demonstration dataset has $N$ trajectories, $\mathcal D=\{\tau_i\}_{i=1}^{N}$. The training objective is to maximise the action log-likelihood under the expert data distribution:
\begin{equation}
    \theta^\star=\arg\max_\theta\ \mathbb E_{(o_t,a_t)\sim\mathcal D}\big[\log\pi_\theta(a_t\mid o_t)\big].
\end{equation}
The policy comes from some parametric class parameterised by $\theta$, which could be, for example, the weights of a VLA policy, $\pi_\theta(a_t\mid o_t,c)$, which maps a visual observation $o_t$ and a language instruction $c$ at time step $t$ to a probability distribution over robot actions $a_t$.

\textbf{Memory problems.}
Long-horizon robotic manipulation requires robots to compose perception, reasoning, and action over extended task sequences; however, not all of them need memory, the ability to retain and reuse historical context across time~\citep{ni2023transformers}. 

\begin{definition}[memory problem for a visuomotor policy]\label{def:long-horizon}
Write $h_{t-k:t}=\{(o_j,a_j)\}_{t-k\le j<t}$ for the last $k$ steps of the history. The \textit{memory length} $l_{\mathrm{mem}}$ of a task is the shortest window from which the expert policy can be realised,
\begin{equation}
l_{\mathrm{mem}} \;=\; \min\, k \quad\text{s.t.}\quad \pi^\star\big(a_t\mid h_t,\,o_t\big) = \pi^\star\big(a_t\mid h_{t-k:t},\,o_t\big)\ \ \forall t.
\label{eq:memory-length}
\end{equation}
A policy that conditions on the last $l$ steps, $\pi_\theta\big(a_t\mid h_{t-l:t},\,o_t\big)$, has \textit{context length} $l_{\mathrm{ctx}}=l$~\citep{ni2023transformers}. A policy faces a \textbf{memory problem} if and only if the task's memory length exceeds the policy's context length, $l_{\mathrm{mem}} > l_{\mathrm{ctx}}$, i.e., the input is insufficient to determine the correct action, and the policy must retain and reuse historical context to resolve the ambiguity.
\end{definition}

\section{Optimal memory analysis} \label{sec:analysis}

\begin{figure}[t]
\centering
\includegraphics[width=0.7\linewidth]{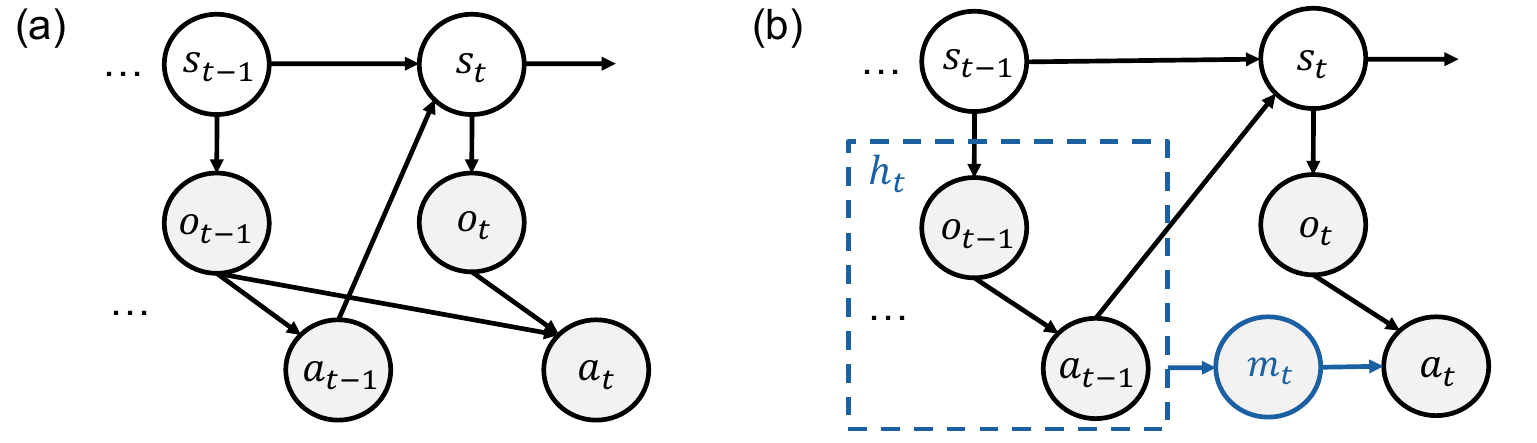}
\caption{Graphical models of (a) imitation learning in a POMDP, where $a_t$ depends on the history $h_t$, and (b) a memory-augmented policy, where $m_t$ makes $a_t$ independent of $h_t$ given $(o_t,m_t)$.}
\label{fig:gm}
\end{figure}

This section derives the objective behind ``what to remember'': the information a memory $m_t=\mathcal M(h_t)$ must carry so that $a_t$ is independent of $h_t$ given $(o_t,m_t)$ (Figure~\ref{fig:gm}(b)):

\begin{proposition}[optimal memory for visuomotor policies]\label{prop:optmem}
The information that the history $h_t$ carries for control beyond the current observation is the conditional mutual information $I\big(a_t;h_t\mid o_t\big) = \mathbb E_{p(h_t,o_t)}[\KL\big(\pi^\star(\cdot\mid o_t,h_t)\,\big\|\,p(\cdot\mid o_t)\big)]$, the expected divergence between the expert policy and the action distribution obtained by marginalising it over the history. Let the memory be any function of the full history, $m_t=\mathcal M(h_t)$, inserted into the graphical model of Figure~\ref{fig:gm}(b). By the chain rule, the memory splits this information into the part it keeps and the part it discards,
\begin{equation}
I\big(a_t;\,h_t\mid o_t\big)
\;=\; I\big(a_t;\,m_t\mid o_t\big)
\;+\;I\big(a_t;\,h_t\mid o_t,m_t\big).
\label{eq:mem-decomp}
\end{equation}
Since the left-hand side does not depend on the memory, maximising $I\big(a_t;\,m_t\mid o_t\big)$ is equivalent to minimising the discarded information $I\big(a_t;\,h_t\mid o_t,m_t\big)$, so the \textbf{optimal memory for visuomotor policies} is 
\begin{equation}
\mathcal M^\star=\arg\max_{\mathcal M}\; I\big(a_t;\,\mathcal M(h_t)\mid o_t\big).
\label{eq:keep-equals-lose}
\end{equation}
The memory $m_t=\mathcal M^\star(h_t)$ is a policy-sufficient statistic, i.e., $\pi^\star(a_t\mid o_t,h_t)=\pi^\star(a_t\mid o_t,m_t)$ for all $(o_t,h_t)$.
\end{proposition}

Other solutions to POMDPs assume richer statistics of the history, for instance a belief that also predicts reward and transition; the more a statistic must carry, the more it demands of the model, the compute, and the memory budget (Appendix~\ref{app:belief}).

The objective gives a unified view of existing memories. Perceptual memory keeps the frames that differ most from the current one, which maximises $H(o_k\mid o_t)$, an upper bound on what a past frame $o_k$ adds since $I(a_t;\,o_k\mid o_t)\le H(o_k\mid o_t)$. 
The bound is loose when the observation changes for reasons unrelated to the action (e.g., a lighting change)~\citep{dunion2023cmid} and when the robot's behaviour is barely visible (e.g., a button press).

To see what can actually be optimised, we unfold $I\big(a_t;\,m_t\mid o_t\big)$:
\begin{align}
I\big(a_t;\,m_t\mid o_t\big)
& = H(a_t\mid o_t)\;-\;H\big(a_t\mid o_t,m_t\big)
&& \text{\small(definition)}\label{eq:mi-entropy}\\[2pt]
&= \mathbb E_{\mathcal D}\big[\log p(a_t\mid o_t,m_t)\big] + H(a_t\mid o_t)
&& \text{\small(expand the residual)}\label{eq:gt_obj_wo_policy}\\[2pt]
&\;\ge\; \mathbb E_{\mathcal D}\big[\log \pi_\theta(a_t\mid o_t,m_t)\big] + H(a_t\mid o_t)
&& \text{\small(variational $\pi_\theta$ for $p$)}\label{eq:vlb}
\end{align}
\Eqref{eq:mi-entropy} and~\Eqref{eq:gt_obj_wo_policy} are defined under the true $p(a_t\mid o_t,m_t)$ of the demonstration distribution, which is unknown and cannot be evaluated directly. 
\Eqref{eq:vlb} replaces $p$ with the policy $\pi_\theta$, giving a lower bound that tightens as $\pi_\theta$ fits the data, while the entropy $H(a_t\mid o_t)$ is fixed by the data. Optimising a mutual-information objective through such a tractable bound has proven effective for generalisable control in reinforcement learning~\citep{yu2024skillaware}. \textbf{\boldmath The objective becomes maximising $\mathbb E_{\mathcal D}[\log\pi_\theta(a_t\mid o_t,m_t)]$ over \emph{two} variables: the policy parameters $\theta$ and the memory $m_t$}:
\begin{equation}
\max_{\mathcal M}\; I\big(a_t;\,m_t\mid o_t\big)
\;\;\ge\;\;
\max_{\mathcal M,\,\theta}\; \mathbb E_{\mathcal D}\big[\log \pi_\theta(a_t\mid o_t,m_t)\big] + H(a_t\mid o_t),
\label{eq:mi-equals-loss}
\end{equation}
where the right-hand side is the negative action loss and a constant. 
In practice $\pi_{0.5}$~\citep{pi05_2025} trains with a flow-matching loss, which serves as a surrogate for the action log-likelihood here (it trains the same conditional action distribution but does not bound its likelihood). Optimising the right-hand side of \Eqref{eq:mi-equals-loss} through that loss opens a design space in which policy and memory are trained together end-to-end, \emph{without} any auxiliary loss or hand-designed rule. Latent memory trained with the action loss, as HAMLET is, already lives in this space, but it optimises the bound through a fixed compression of every step, and so loses the fidelity of the pretrained tokens.

\section{Divide-and-Remember}\label{sec:method}

This section describes how we optimise $I(a_t;\,m_t\mid o_t)$ in practice. We keep the memory in the pretrained token space of the VLA, so that $m_t$ is a subset of the history tokens and a kept fact can be read back exactly; memory optimisation then becomes a discrete problem, learning the mapping that selects, from the history tokens, the subset that maximises $I(a_t;\,m_t\mid o_t)$ under a budget of $K$ tokens. Realising such an $\mathcal M$ raises two challenges: (i) $m_t=\mathcal M(h_t)$ is a function of the whole $h_t$, which grows without bound, yet $\mathcal M$ must stay \textit{lightweight}, and (ii) $\mathcal M$ must bridge an \textit{offline--online mismatch}: during training the whole history of an episode is available at once from $\mathcal D$, whereas at deployment it arrives one step at a time. Section~\ref{sec:method_submodular} casts memory as budgeted token selection and introduces \ours that meets both challenges; Section~\ref{sec:method_selector} describes the selector and its training.

\begin{figure}[t]
\centering
\includegraphics[width=\linewidth]{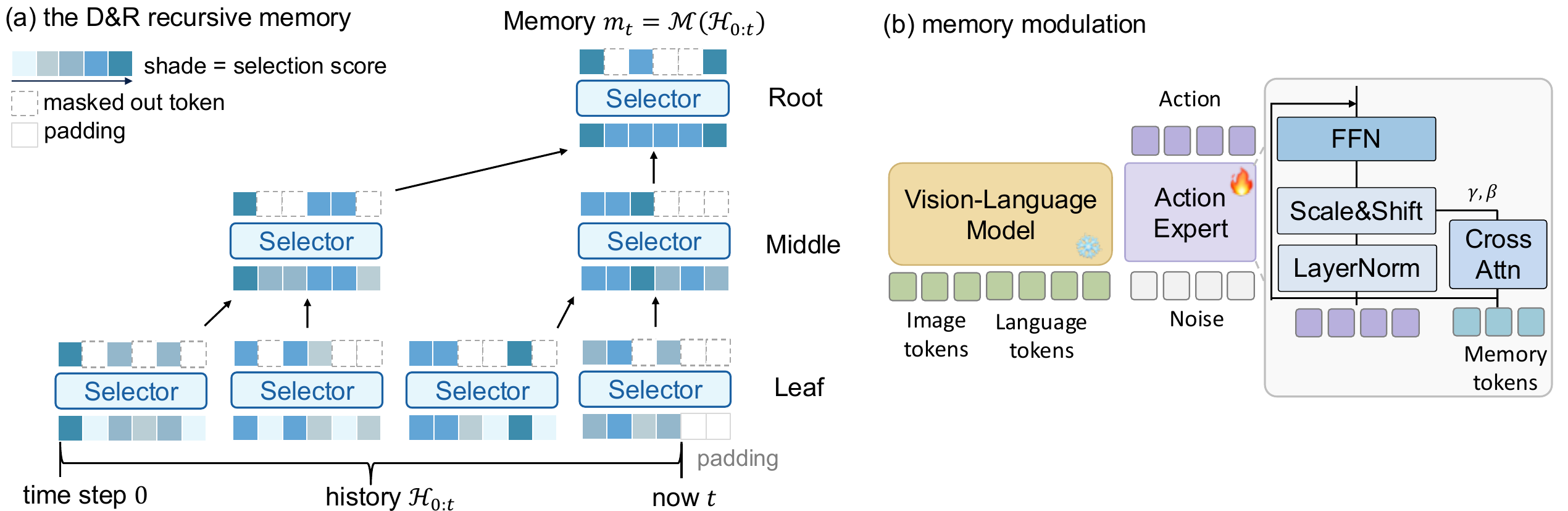}
\caption{Overview of \ours. (a) At time step $t$, the selection over the full history $\mathcal{H}_t$ is divided recursively into subproblems of top-$K$ selection over $2K$ tokens, each solved by the same lightweight selector $f_\psi$; the kept tokens are merged pairwise and re-selected level by level, and at the root the selected tokens are passed to the policy through the attention mask. Tokens keep their temporal order from left to right at every level. 
(b) \ours and all baselines modulate the memory into the action expert through an adaptive LayerNorm, where action features cross-attend to the memory tokens.}
\label{fig:tree}
\end{figure}

\subsection{Memory as budgeted token selection}\label{sec:method_submodular}

Let $\mathcal H_t$ denote the set of all history tokens available at time step $t$. Index these tokens in temporal order, and let $\mathcal H_{i:j}\subseteq\mathcal H_t$ denote the tokens at positions $i$ through $j$. A memory is a mapping $\mathcal M$ from the history tokens to a subset of at most $K$ of them, $m_t=\mathcal M(\mathcal H_t)\subseteq\mathcal H_t$, $|m_t|\le K$; let $\mathfrak M_K$ denote the set of such mappings. Writing $I(\mathcal M)$ for $I(a_t;\,\mathcal M(\mathcal H_t)\mid o_t)$, the memory problem is
\begin{equation}
\mathcal M^\star=\argmax_{\mathcal M\in\mathfrak M_K}\ I\big(a_t;\,\mathcal M(\mathcal H_t)\mid o_t\big).
\label{eq:memory-submodular}
\end{equation}
The difficulty is that
$I$ is not available in closed form, so we settle for an approximate solution: our approach uses a parametric family $\mathcal M_\psi$ built from a selector learned end-to-end, optimised over $\psi$. Moreover, $\mathcal H_t$ grows with the history while a selector has a fixed input size. {We fix the selector's input at $2K$ tokens and build $\mathcal M_\psi$ recursively: the selection over the full history is composed from subproblems of top-$K$ selection over $2K$ tokens, so one fixed-size, lightweight selector serves an unbounded history.} 

\begin{figure}[t]
\noindent
\begin{minipage}[t]{0.52\textwidth}
\vspace{0pt}
\vspace{-\intextsep}
\begin{algorithm}[H]
\caption{Divide-and-Remember}
\label{alg:dnr}
\small
\begin{algorithmic}[1]
\Require span $\mathcal H_{i:j}$; budget $K$; selector $f_\psi$
\Ensure $\mathcal M_\psi(\mathcal H_{i:j})\subseteq\mathcal H_{i:j}$, at most $K$ tokens
\If{$|\mathcal H_{i:j}|\le K$}
    \State \Return $\mathcal H_{i:j}$
\EndIf
\State $\ell\gets\max\{\,\ell'\ge 0 \;:\; 2^{\ell'}K<|\mathcal H_{i:j}|\,\}$
\State $q\gets i+2^{\ell}K-1$ \Comment{left-aligned split}
\State \Return $f_\psi\big(\mathcal M_\psi(\mathcal H_{i:q})\cup\mathcal M_\psi(\mathcal H_{q+1:j})\big)$
\end{algorithmic}
\end{algorithm}
\end{minipage}\hfill
\begin{minipage}[t]{0.45\textwidth}
\vspace{0pt}
\vspace{-\intextsep}
\begin{figure}[H]
\centering
\setlength{\abovecaptionskip}{2pt}
\includegraphics[width=\linewidth]{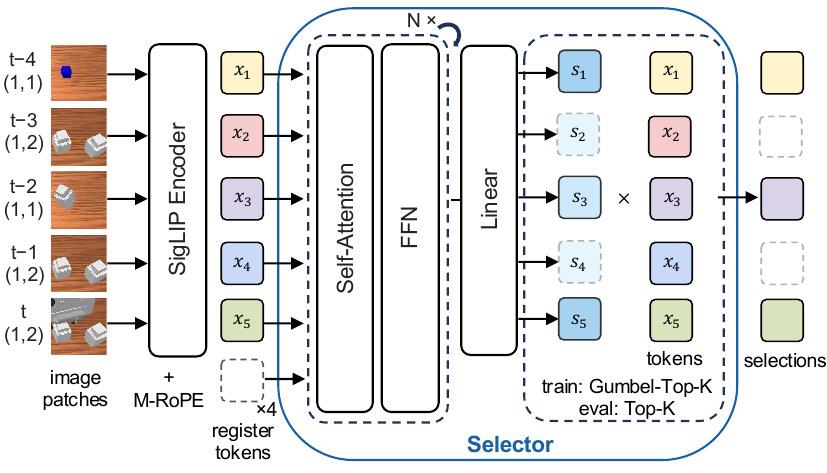}
\caption{The selector $f_\psi$.}
\label{fig:selector}
\end{figure}
\end{minipage}
\end{figure}

We instantiate this approach in a method we call \textbf{Divide-and-Remember} (\ours). Writing $f_\psi$ for the learned selector that maps $2K$ tokens to their top-$K$ (Section~\ref{sec:method_selector}), the memory of a span $\mathcal H_{i:j}$ is
\begin{equation}
    \mathcal M_\psi(\mathcal H_{i:j})=
    \begin{cases}
        \mathcal H_{i:j}, & \text{if }|\mathcal H_{i:j}|\le K \\[4pt]
        f_\psi\big(\mathcal M_\psi(\mathcal H_{i:q})\cup\mathcal M_\psi(\mathcal H_{q+1:j})\big), & \text{otherwise}
    \end{cases}
    \label{eq:memory-recursion}
\end{equation}
where the split is left-aligned, $q=i+2^{\ell}K-1$ with $2^{\ell}K$ the largest block size that fits strictly inside $\mathcal H_{i:j}$, and the memory at step $t$ is $m_t=\mathcal M_\psi(\mathcal H_t)$. Because $\mathcal M_\psi$ composes over token spans rather than over a fixed window, {training and deployment build the same \ours tree for the same prefix $\mathcal H_t$ of an episode, so $\mathcal M_\psi$ is free of the offline--online mismatch}. Figure~\ref{fig:tree}(a) illustrates the construction and Algorithm~\ref{alg:dnr} gives the pseudocode.

The recursion filters tokens hierarchically from the leaves to the root: only tokens kept at one level are re-scored at the next, so computation is spent only on the tokens that survive. Higher levels thus hold increasingly important and sparse information, while the least important, such as background, is dropped at the leaves.

\subsection{Learning the selector}\label{sec:method_selector}

The selector $f_\psi$ takes $2K$ candidate history tokens $\{x_i\}_{i=1}^{2K}$ in temporal order, encoded by the frozen SigLIP encoder, and returns the top-$K$ of them as the memory of that slice. To preserve the spatiotemporal structure of the tokens, we adopt multimodal rotary position embedding (M-RoPE)~\citep{wang2024qwen2vl} over time, height, and width. A transformer of $N$ self-attention and FFN blocks then scores each token, and a linear head outputs a selection score $s_i$. The memory is
\begin{align}
\hat m &= \{\,g_i\,x_i \mid i\in\mathrm{TopK}(\{s_j\}_{j=1}^{2K},K)\,\},\\
g_i &= \operatorname{sg}\big(g_i^{\text{hard}}-g_i^{\text{soft}}\big)+g_i^{\text{soft}},
\end{align}
where $g_i^{\text{hard}}=\mathbb 1[s_i>\rho]$ is the hard selection, $g_i^{\text{soft}}=\sigma((s_i-\rho)/\tau)$ its sigmoid relaxation with temperature $\tau$, $\operatorname{sg}$ stops the gradient, and $\rho$ is the threshold, the detached midpoint between the $K$-th and $(K{+}1)$-th largest scores. Selected tokens thus have $g_i=1$ in the forward pass and receive gradient through $g_i^{\text{soft}}$ in the backward pass~\citep{rao2021dynamicvit}. During training, Gumbel noise is added to $s_i$ before the top-$K$ for exploration~\citep{kool2019gumbel}; at the root, $g_i$ masks the attention over the $2K$ candidates, and the masked candidates $\hat m$ are passed to the policy.
A single selector $f_\psi$ is shared across all nodes to keep training and inference efficient, and its only training signal is the flow-matching action loss of the policy that reads the selected tokens, the variational lower bound of $I(a_t;\,m_t\mid o_t)$ in \Eqref{eq:mi-equals-loss}.

\section{Experiments}\label{sec:experiments}

Our experiments aim to answer whether \ours improves consistently across tasks with different memory demands, and where each memory method helps and why.

\begin{figure}[t]
\centering
\includegraphics[width=0.9\linewidth]{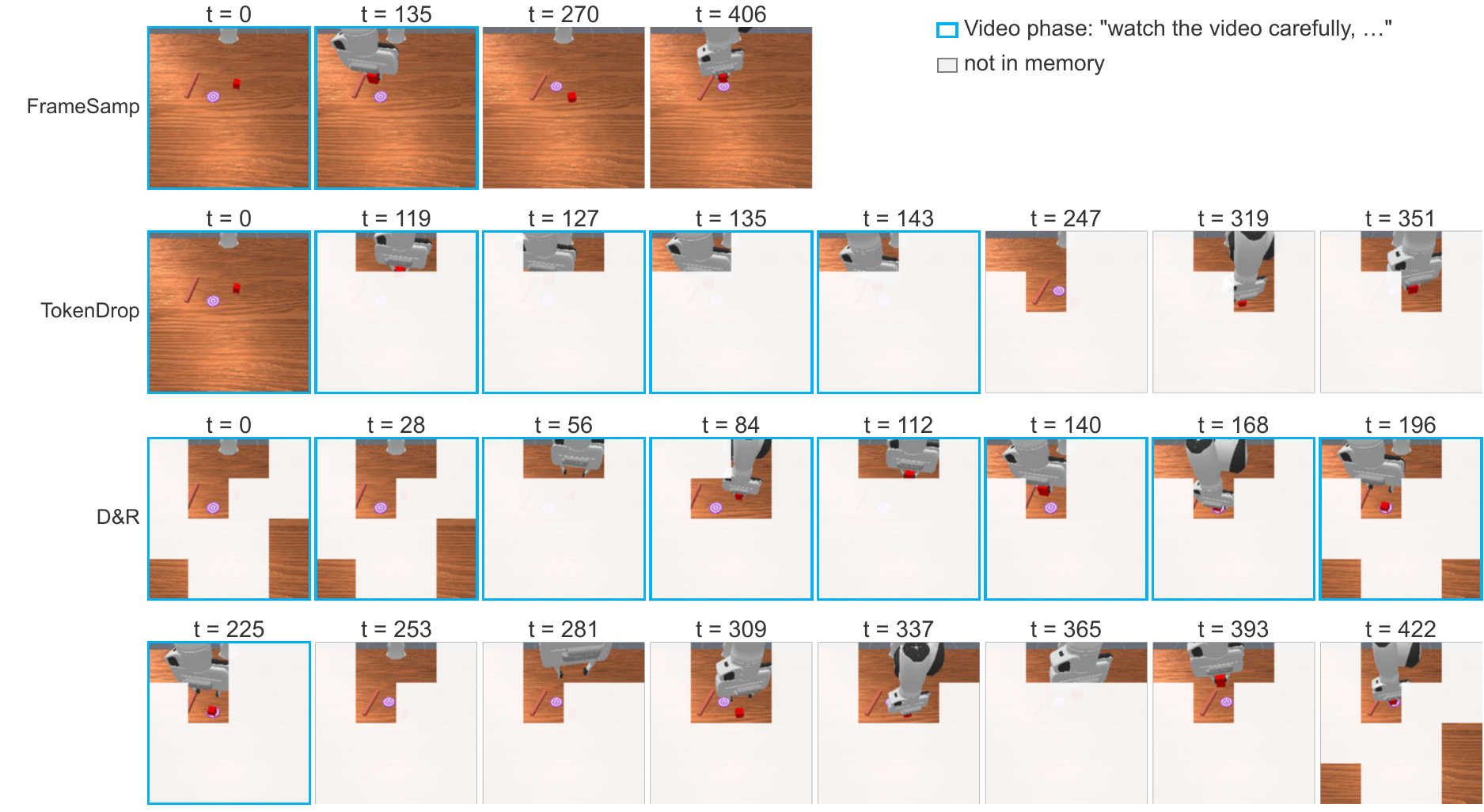}
\caption{Memory at step $422$ of a MoveCube episode under a $64$-token budget, with the task goal ``watch the video carefully, then move the cube to the target in the same manner as before''. FrameSamp keeps $4$ evenly spaced frames. TokenDrop keeps the first frame and spends the rest of the budget on the patches whose pixels changed most. \ours selects $64$ patches recursively, $35$ of them from the video phase. Kept patches are shown, white ones are not in memory, and light blue borders mark the video phase. All tasks are in Appendix~\ref{app:memory_vis}.}
\label{fig:memory_vis}
\end{figure}

\subsection{Experimental setup}
We evaluate on RoboMME~\citep{dai2026robomme}, a benchmark of $16$ long-horizon manipulation tasks that each place a different demand on memory. We group them by their six functional characteristics: Motion-Centric (tracking a trajectory), Time-Sensitive (acting at the correct moment), Short- and Long-Horizon Video (grounding objects from a short or an extended video), Dynamic Scene-Change (adapting to a scene that changes during execution), and Event-Salient (progress marked by discrete events). 
All models are trained on the RoboMME demonstrations of all $16$ tasks in a multi-task setting: the SigLIP encoder is frozen, and the Gemma VLM, the action expert, and the parameters a memory adds (for \ours, the selector) are fine-tuned with the flow-matching action loss. The policy predicts a chunk of $20$ actions and executes the first $16$. Every model is trained for $80$k steps with a batch size of $64$ (TTT uses $16$); implementation details and results by the benchmark's four suites are in Appendix~\ref{app:exp}.

\textbf{Baselines.}
To answer what information the memory should carry, we compare against the three memory families: perceptual memory, TokenDrop and FrameSamp~\citep{dai2026robomme}; latent memory, HAMLET~\citep{koo2026hamlet} and RB-VLA~\citep{bagaria2026rbvla}; and parametric memory, TTT~\citep{sun2024ttt}. All memory methods build on the $\pi_{0.5}$ backbone and are modulated into the action expert in the same way (Figure~\ref{fig:tree}(b)). Our code, including the $\pi_{0.5}$ re-implementations of HAMLET and RB-VLA, with checkpoints and more results, are at \href{https://dnr-memory.github.io/}{dnr-memory.github.io}.

\textbf{Evaluation protocol.} All memory methods are given the same budget of $64$ tokens and differ only in what fills it. Each model is evaluated on all $16$ tasks with $50$ episodes per task ($800$ in total), on environment seeds disjoint from training and a maximum horizon of $1{,}300$ steps. Results are averaged over the last three checkpoints and three random seeds (nine runs in total).

\subsection{Experimental results}
\begin{table}[t]
\centering
\caption{
Per-task success rate (\%) on RoboMME~\citep{dai2026robomme}, tasks grouped by functional characteristic (the same results by suite are in Table~\ref{tab:full-results}). All memories are built on $\pi_{0.5}$ with a budget of $64$ tokens and differ in what they store: raw patch tokens kept by frame sampling (FrameSamp) or token dropping (TokenDrop); learned moment tokens (HAMLET) or a recursive belief (RB-VLA); or fast weights (TTT). Each number is the mean over the last three checkpoints and three seeds; standard deviations are in Table~\ref{tab:full-results}. Avg is the mean over the tasks of a characteristic; \textbf{bold} is the highest success rate in each column.}
\label{tab:robomme-taskchar}
\setlength{\tabcolsep}{3pt}\renewcommand{\arraystretch}{1.15}
\resizebox{\linewidth}{!}{%
\begin{tabular}{@{}l l >{\columncolor{black!8}}c >{\columncolor{black!8}}c >{\columncolor{black!8}}c >{\columncolor{black!8}}c >{\columncolor{black!8}}c c >{\columncolor{black!8}}c >{\columncolor{black!8}}c >{\columncolor{black!8}}c >{\columncolor{black!8}}c c c c >{\columncolor{black!8}}c >{\columncolor{black!8}}c >{\columncolor{black!8}}c >{\columncolor{black!8}}c c c c c@{}}
\toprule
 &  & \multicolumn{5}{>{\columncolor{black!8}}c}{\textbf{Motion-Centric}} & \multicolumn{1}{c}{\textbf{Time-Sensitive}} & \multicolumn{4}{>{\columncolor{black!8}}c}{\textbf{Short-Horizon Video}} & \multicolumn{3}{c}{\textbf{Long-Horizon Video}} & \multicolumn{4}{>{\columncolor{black!8}}c}{\textbf{Dynamic Scene-Change}} & \multicolumn{4}{c}{\textbf{Event-Salient}} \\
\cmidrule(lr){3-7}\cmidrule(lr){8-8}\cmidrule(lr){9-12}\cmidrule(lr){13-15}\cmidrule(lr){16-19}\cmidrule(lr){20-23}
Type & Method
& \makecell{Pattern\\Lock} & \makecell{Route\\Stick} & \makecell{Swing\\Xtimes} & \makecell{Insert\\Peg} & Avg
& \makecell{Stop\\Cube}
& \makecell{Video\\Umsk} & \makecell{Video\\Repick} & \makecell{Video\\UmskS} & Avg
& \makecell{Video\\PlcBtn} & \makecell{Video\\PlcOrd} & Avg
& \makecell{Button\\Umsk} & \makecell{Button\\UmskS} & \makecell{Pick\\HighL} & Avg
& \makecell{Pick\\Xtimes} & \makecell{Bin\\Fill} & \makecell{Move\\Cube} & Avg \\
\midrule
No memory & $\pi_{0.5}$ & 2.89 & 4.67 & 35.56 & 1.56 & 11.17 & 6.67 & 20.44 & 0.44 & 18.67 & 13.18 & 31.11 & 25.78 & 28.45 & 22.22 & 6.67 & 11.33 & 13.41 & 42.89 & 30.00 & 26.00 & 32.96 \\
\midrule
\multirow{2}{*}{Perceptual} & FrameSamp & 15.56 & 17.78 & 88.45 & 2.22 & 31.00 & 27.34 & 30.00 & 14.22 & 20.89 & 21.70 & 26.89 & 25.56 & 26.23 & 6.67 & 5.11 & 18.67 & 10.15 & 60.44 & 34.44 & 52.66 & 49.18 \\
 & TokenDrop & 9.56 & 8.89 & 36.67 & 2.44 & 14.39 & 5.78 & 28.67 & 1.56 & 16.67 & 15.63 & 26.67 & 19.56 & 23.12 & 16.67 & 6.44 & 14.00 & 12.37 & 39.78 & 27.78 & 24.44 & 30.67 \\
\midrule
\multirow{2}{*}{Latent} & HAMLET & 7.33 & 9.56 & 81.11 & 1.11 & 24.78 & \textbf{39.33} & 27.78 & 20.67 & 25.11 & 24.52 & 27.78 & 21.78 & 24.78 & 26.67 & \textbf{21.78} & 26.00 & 24.82 & \textbf{84.67} & 34.89 & \textbf{60.00} & \textbf{59.85} \\
 & RB-VLA & 0.67 & 0.20 & 86.00 & \textbf{3.33} & 22.55 & 10.67 & 30.00 & 26.00 & 23.33 & 26.44 & 24.67 & 24.00 & 24.34 & \textbf{28.00} & 20.00 & \textbf{26.67} & \textbf{24.89} & 79.33 & \textbf{36.67} & 36.67 & 50.89 \\
\midrule
Parametric & TTT & 4.67 & 7.11 & 41.56 & 0.89 & 13.56 & 2.89 & 28.67 & 4.44 & 19.78 & 17.63 & 20.44 & 17.11 & 18.77 & 15.11 & 3.33 & 12.44 & 10.29 & 42.67 & 28.22 & 24.67 & 31.85 \\
\midrule
Ours & \ours & \textbf{32.00} & \textbf{59.33} & \textbf{95.33} & \textbf{3.33} & \textbf{47.50} & 31.33 & \textbf{32.67} & \textbf{26.67} & \textbf{27.33} & \textbf{28.89} & \textbf{34.00} & \textbf{29.33} & \textbf{31.67} & 26.00 & 21.33 & 24.00 & 23.78 & 82.67 & 35.33 & 56.67 & 58.22 \\
\bottomrule
\end{tabular}}
\end{table}

\paragraph{Main results.}
Overall, \ours reaches the highest average success rate over the four suites ($38.6\%$, against $32.2\%$ for HAMLET, $27.9\%$ for FrameSamp, and $17.9\%$ for the memoryless $\pi_{0.5}$; Table~\ref{tab:full-results}). The memory visualisations show that \ours retrieves the key fact: in Figure~\ref{fig:memory_vis}, for example, it keeps the grasp pose from the MoveCube video at the start of the episode, more than $300$ steps earlier. Latent memory collapses on Imitation ($19.5\%$ for HAMLET and $10.2\%$ for RB-VLA, which scores $0.0\%$ on RouteStick): a compressed state was unable to adequately preserve the required spatial detail. More results are in Appendix~\ref{app:sim-results}.

To understand where each memory helps, we report the results by the functional characteristics of the tasks (Table~\ref{tab:robomme-taskchar}) rather than by the four suites, which name cognitive memory types. We categorise the historical key fact into frame-like, one past frame that the current observation points to (Motion-Centric, Short- and Long-Horizon Video), and state-like, the running state of a process that unfolds during the episode, such as elapsed time, a changing scene, or a count of events (Time-Sensitive, Dynamic Scene-Change, Event-Salient); the two kinds separate the memory methods clearly. The key observations are: (i) \ours gains the most on \emph{frame-like facts}: $47.5\%$ on Motion-Centric against $31.0\%$ for FrameSamp and $24.8\%$ for HAMLET, and on Long-Horizon Video it is the only memory above the memoryless policy ($31.7\%$ against $28.5\%$). (ii) Latent memory is competitive on \emph{state-like facts}: HAMLET leads Time-Sensitive ($39.3\%$ against $31.3\%$) and ties \ours on Dynamic Scene-Change and Event-Salient ($24.8$ and $59.9$ against $23.8$ and $58.2$), where RB-VLA also ties on the former ($24.9$). A latent memory updated at every step tracks a running state by construction, whereas \ours has to reconstruct it from snapshots. For example, in BinFill, it learns to keep one snapshot of each cube dropped into the bin (Figure~\ref{fig:memory_vis_BinFill}). %

On InsertPeg, every method stays at or below $3.3\%$, against $1.6\%$ for the memoryless $\pi_{0.5}$. The rollout in Figure~\ref{fig:memory_vis_InsertPeg} reaches the correct end of the peg and the correct side of the box, so the failure lies in the precision of the insertion rather than in memory.

\begin{figure}[t]
\centering
\includegraphics[width=\linewidth]{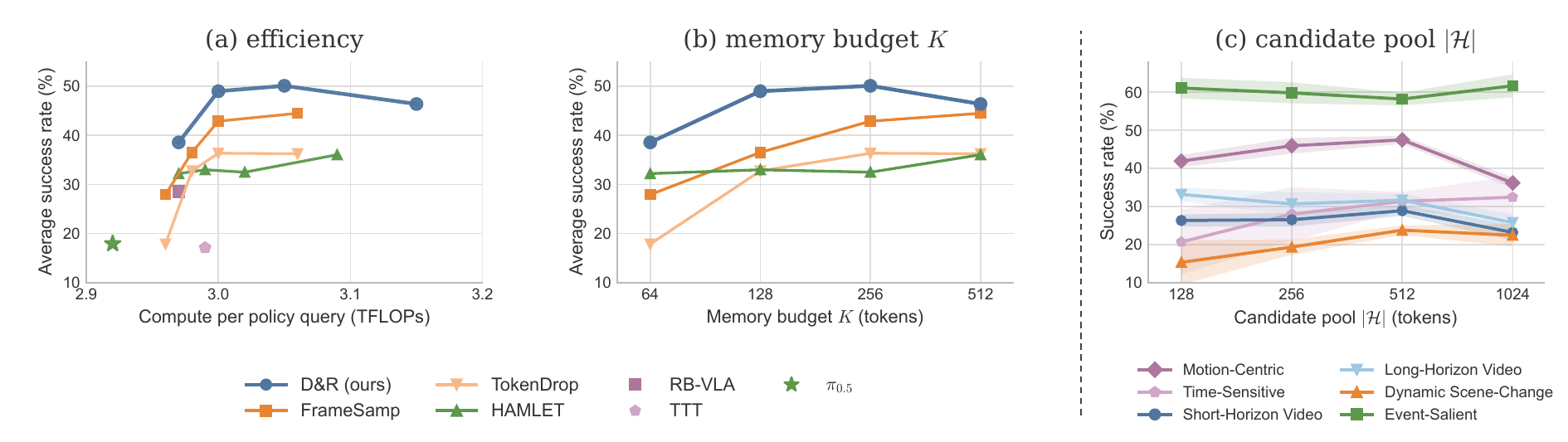}
\caption{(a) Average success rate against the TFLOPs of one policy forward pass; each line sweeps the memory budget $K$ from $64$ to $512$ (Table~\ref{tab:memory_budget_ablation}). (b) Average success rates against $K$. (c) \ours per task characteristic~\citep{dai2026robomme} as the candidate pool $|\mathcal{H}|$ grows from $128$ to $1024$ tokens at $K=64$; mean $\pm$ std over runs (Table~\ref{tab:dnr-pool-taskchar}).}
\label{fig:analysis}
\end{figure}

\textbf{Efficiency against performance.}
Memory buys sufficiency with computation. Varying the budget from $64$ to $512$ tokens and measuring cost as the TFLOPs of one forward pass at the average episode length (Figure~\ref{fig:analysis}(a)), \ours reaches the highest success at essentially the same cost per query. RB-VLA, by contrast, updates its belief recurrently at every step, so training cannot be parallelised over time and takes about $10$ days for $80$k steps, whereas \ours takes about $20$ hours (Appendix~\ref{app:exp}).

\begin{wraptable}{r}{0.42\textwidth}
\vspace{-\intextsep}
\centering\scriptsize
\caption{Real-world results.}
\label{tab:realworld}
\setlength{\tabcolsep}{3pt}\renewcommand{\arraystretch}{1.05}
\begin{tabular}{@{}l cccc c@{}}
\toprule
Method & \makecell{Put\\Bottles} & \makecell{Track\\Cube} & \makecell{Repick\\Cube} & \makecell{Draw\\Pattern} & Total \\
\midrule
$\pi_{0.5}$ & 0/10 & 2/10 & 0/10 & 1/10 & 3/40 \\
FrameSamp   & 4/10 & 8/10 & 6/10 & 4/10 & 22/40 \\
\ours       & \textbf{9/10} & \textbf{9/10} & \textbf{9/10} & \textbf{8/10} & \textbf{35/40} \\
\bottomrule
\end{tabular}
\end{wraptable}

\textbf{Real-world robot manipulation.} We test whether memory keeps a real robot reliable over several long-run tasks, with a person intervening mid-task and under perception noise~\citep{dey2026alliance}. On four tasks, PutBottles, TrackCube, RepickCube and DrawPattern, \ours succeeds in $35$ of $40$ trials, against $22$ for FrameSamp and $3$ for $\pi_{0.5}$ (Table~\ref{tab:realworld}), with the largest margin on PutBottles, the longest task at up to $4$ minutes. FrameSamp fails when its memory holds the wrong frame, and errors accumulate over time, with most failures on the second cube of RepickCube and the third move of DrawPattern. Without memory, $\pi_{0.5}$ falls back on a fixed behaviour, starting eight of its nine failed DrawPattern trials with the same leftward move whatever the video showed. Details are in Appendix~\ref{app:real-robot}.

\subsection{Ablation} \label{sec:exp_analysis}

\textbf{Candidate pool size.}
\ours does not need to read the full history: the selector works on a candidate pool of $|\mathcal{H}|$ tokens, $|\mathcal{H}|/16$ frames sampled uniformly from the history with $16$ pooled patch tokens each, which keeps its cost independent of the growing history length. We sweep $|\mathcal{H}|$ from $128$ to $1024$ at the same budget $K=64$ (Tables~\ref{tab:dnr-pool} and~\ref{tab:dnr-pool-taskchar}). The key observations are: state-like facts grow with the pool ($20.7\%$ to $32.4\%$ on Time-Sensitive), since covering more of the time axis helps track the process, whereas frame-like facts need specific tokens retrieved at full fidelity, and a larger pool hides them among more candidates, so the selector is harder to learn; Motion-Centric peaks at $512$ ($47.5\%$) and drops at $1024$ ($36.2\%$).

\textbf{Which sufficient statistic to learn.}
To ablate how the training objective (the sufficient statistic) shapes what the memory builds up, we add \ours{}+pred, which trains \ours with RB-VLA's world-model loss on top of the action loss (Appendix~\ref{app:variants}); \ours, \ours{}+pred, and RB-VLA then span policy sufficiency, sufficiency plus prediction, and a fully predictive belief. 
At a pool of $|\mathcal{H}|=256$ (Table~\ref{tab:dnr-ablations}), the prediction loss brings no gain: asking the memory to predict the future spends capacity that the action does not buy back (Section~\ref{sec:analysis}). 

\textbf{Historical token types.}
Adding the instruction's text tokens to the candidate pool ($+$text in Table~\ref{tab:dnr-ablations}) costs $6.6$ points overall and $14.5$ on imitation. Our main experiments therefore use patch tokens only.
Action tokens are excluded from the candidates because of causal confusion: the action-conditioned policy might learn to ``cheat'' by simply outputting the previous action, which is often highly (yet spuriously) correlated with the next action.

\textbf{Effect of the memory budget.}
Figure~\ref{fig:analysis}(b) plots the average success rate as the memory budget grows from $64$ to $512$ tokens. \ours leads at every budget, but its curve flattens at $256$ tokens ($50.1\%$) and dips at $512$ ($46.4\%$), which we attribute to convergence: a larger budget needs more training steps, and at the $80$k steps reported here the $512$-token model is still improving.

\section{Conclusion, limitations, and future work}
In this paper, we proposed Divide-and-Remember (\ours), a recursive memory method that composes a single lightweight selector over the history to maximise the conditional mutual information $I(a_t;\,m_t\mid o_t)$ between the action and the memory in the pretrained token space.
Experimental results showed that \ours achieved state-of-the-art success rates across different memory demands, and that the gain transferred to a real robot.

On state-like facts \ours only ties latent memory, and on StopCube and ButtonUnmask it falls behind, since \ours stores and uses history tokens directly, whereas a latent memory can reformulate or aggregate knowledge in a more compact way. A potential next step is to train the memory to re-value the selected tokens so as to store information more compactly, which could extend its capacity. 

To keep the comparison fair, every memory method was trained in the same multi-task setting without any task-specific information. As a result, the memory modules may converge to acquiring the knowledge shared across tasks, thereby leading to redundancy. There is still design space for a memory method that can enhance the parameter efficiency and also retain specialisation by focusing on distinctive aspects.

\clearpage
\section*{AI use statement}

In this work, we used generative AI tools for editing and checking the manuscript, for literature search, and for refining code. We have not used generative AI tools for research ideation, for generating the method implementation, or for producing experimental data or results; the remaining categories are not applicable to this work. We have reviewed all AI-assisted work, every derivation was checked by the authors, and all text was read and edited by the authors. We take responsibility for the final content of this work, including text, claims or artifacts produced with the aid of generative AI.

\bibliography{iclr2027_conference}
\bibliographystyle{iclr2027_conference}

\clearpage
\appendix
\raggedbottom %

\section{From first principles of POMDP theory}\label{app:belief}

\begin{figure}[t]
\centering
\includegraphics[width=\linewidth]{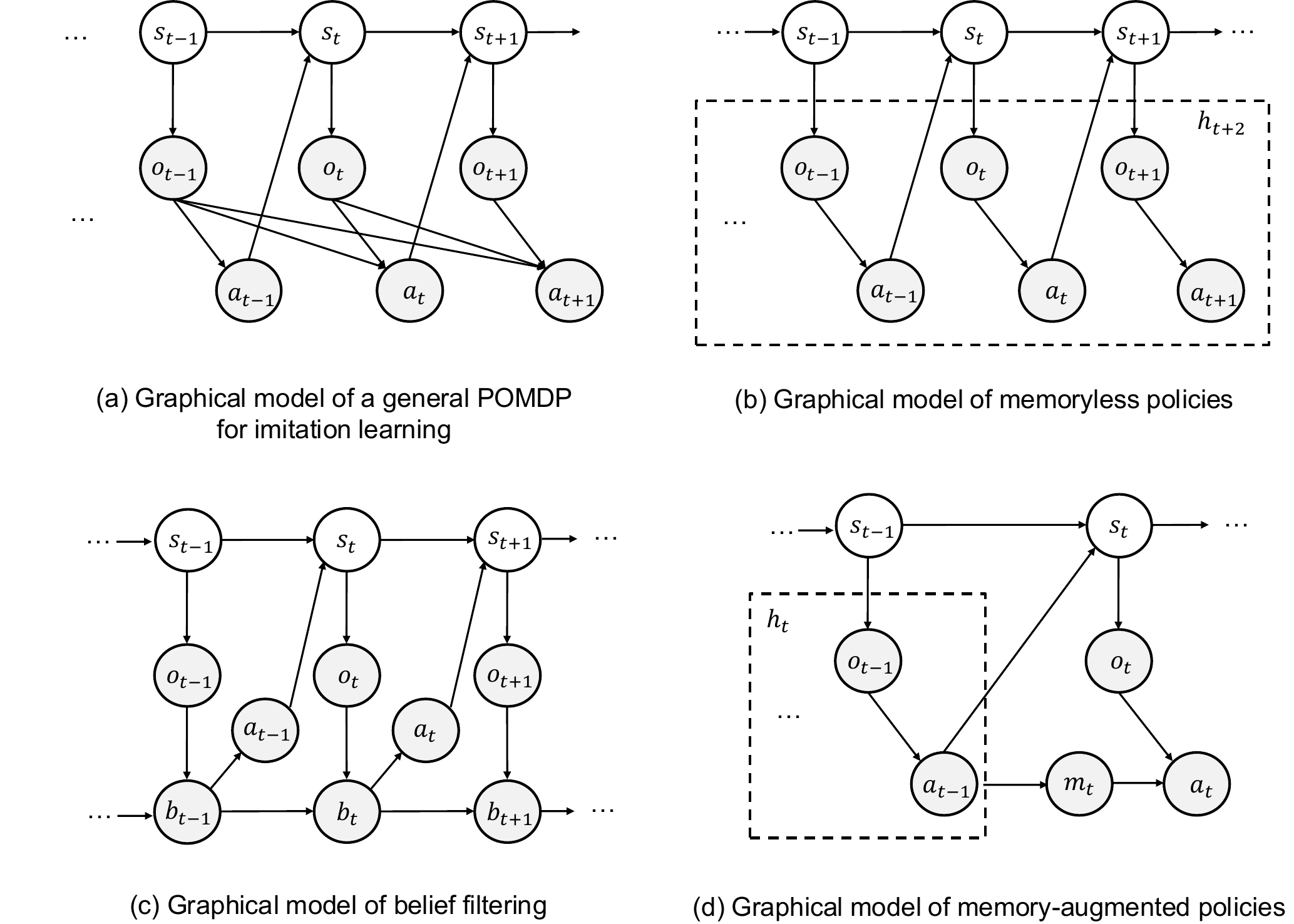}
\caption{\textbf{Graphical models of imitation learning in a POMDP}. 
(a) A general policy for the POMDP inputs the current observation and the full history, $a_t\sim\pi(\cdot\mid o_t,h_t)$, but the history grows without bound.
(b) A memoryless policy $a_t\sim\pi(\cdot\mid o_t)$ inputs only the current observation, which causes the memory problem in Figure~\ref{fig:intro}(a).
(c) A belief policy $a_t\sim\pi(\cdot\mid b_t)$ inputs the belief $b_t(s)=P(s_t=s\mid h_t,o_t)$, the posterior over the current state and a sufficient statistic of the history.
(d) A memory-augmented policy $a_t\sim\pi(\cdot\mid o_t,m_t)$ inputs a memory $m_t=\mathcal M(h_t)$ and the current observation. Here $m_t$ need only be sufficient to reproduce the expert's action (Proposition~\ref{prop:optmem}), and the optimal memory carries the most action-relevant information $I\big(a_t;\,m_t\mid o_t\big)$. Shaded circles are observed in the demonstrations, white circles are hidden.}
\label{fig:pomdp}
\end{figure}

Existing memory methods either select part of the history by an inductive bias (frame novelty, a hand-designed rule, or a verifier), or learn the memory through an auxiliary loss, e.g., predicting the next state. Both report gains, which leaves open how to choose between them, so we return to the first principles of partially observable Markov decision processes (POMDPs), in which the agent receives only partial or noisy observations of the state before it acts (Figure~\ref{fig:pomdp}(a)).

With $h_t=\{(o_k,a_k)\}_{k<t}$ the observations and actions up to time step $t$, one solution is to condition the policy on the whole history, $\pi(\cdot\mid o_t,h_t)$, by solving the associated dynamic decision network~\citep{poole2023artificial}. However, the number of $t$-step histories, and hence the size of such a policy, grows exponentially in $t$, so this works only for short histories. A common alternative is a compact statistic of the history. The belief state, the distribution over states given the history (Figure~\ref{fig:pomdp}(c)), is a sufficient statistic and turns the POMDP into an MDP over beliefs~\citep{kaelbling1998pomdp} (Section~\ref{sec:belief_filtering}), but its dimension grows with the number of states. We therefore seek a representation of the history that scales to the high-dimensional observations and large models of VLAs (Section~\ref{sec:optimal_memory}).

\subsection{Belief} \label{sec:belief_filtering}

\paragraph{Belief is a sufficient statistic.}
Recall that $h_t=\{(o_k,a_k)\}_{k<t}$ contains the observation--action pairs before time $t$, while $o_t$ is the current observation. The belief state $b_t$~\citep{kaelbling1998pomdp} specifies the probability of the agent being in each state given all information available before choosing $a_t$, $b_t(s)=P(s_t=s\mid h_t,o_t)$.

\begin{definition}[sufficient history representation]\label{def:sufficient-history}
A statistic $x_t=\phi(h_t,o_t)$ of the available history and current observation is \textit{sufficient} for the state if
\begin{equation}
P\big(s_t=s\mid h_t,o_t\big)=P\big(s_t=s\mid x_t\big)\qquad\forall s\in\mathcal S,\ \forall (h_t,o_t),
\label{eq:sufficient-history}
\end{equation}
i.e., $x_t$ determines the posterior over $s_t$~\citep{kaelbling1998pomdp}.
\end{definition}

The belief state satisfies~\Eqref{eq:sufficient-history} by construction, so it is a sufficient statistic of the available history and current observation~\citep{albrecht2016causality}. Unfortunately, the dimension of a belief state grows linearly with the number of states, so for a large (e.g.\ continuous) state space we want to learn a more compact representation. 

\citet{kaelbling1998pomdp} show that a belief state can be computed recursively,
\begin{align}
b_{t+1}(s_{t+1})&=P\big(s_{t+1}\mid h_{t+1},o_{t+1}\big) \notag\\
&=P\big(s_{t+1}\mid h_t,o_t,a_t,o_{t+1}\big) \notag\\
&\propto P\big(o_{t+1}\mid s_{t+1}\big)\,P\big(s_{t+1}\mid h_t,o_t,a_t\big) \notag\\
&=P\big(o_{t+1}\mid s_{t+1}\big)\sum_{s_t}P\big(s_{t+1}\mid s_t,a_t\big)\,P\big(s_t\mid h_t,o_t\big),
\label{eq:belief-recursion}
\end{align}
where the last line uses the Markov property of the dynamics and the definition of $b_t$. The recursion shows that $b_{t+1}$ depends on the past only through $b_t$, $a_t$ and $o_{t+1}$. More generally, the state posterior represented by any sufficient $x_t$ can be updated from $(x_t,a_t,o_{t+1})$, as in a recurrent neural network.

Normalising~\Eqref{eq:belief-recursion} gives the \textbf{belief filtering} update~\citep{albrecht2016causality}: for discrete hidden states,
\begin{equation}
    b_{t+1}(s_{t+1})=\frac{p(o_{t+1}\mid s_{t+1})\sum_{s_t\in\mathcal{S}} p(s_{t+1}\mid s_t,a_t)\,b_t(s_t)}{\eta}, \label{eq:belief_update}
\end{equation}
and for continuous hidden states,
\begin{equation}
b_{t+1}(s_{t+1})=\frac{p(o_{t+1}\mid s_{t+1}) \int p(s_{t+1}\mid s_t,a_t)\,b_t(s_t)\,\mathrm{d}s_t}{\eta},
\label{eq:belief_update_cont}
\end{equation}
where $\eta$ normalises the numerator over $s_{t+1}\in\mathcal S$. The two factors play different roles: the transition $p(s_{t+1}\mid s_t,a_t)$ propagates the belief through the dynamics alone and carries no new evidence, while the observation function $p(o_{t+1}\mid s_{t+1})$ reweights that prediction by the incoming observation.

\paragraph{Bisimulation for measuring the sufficiency of a belief state.}
Bisimulation metrics measure state similarity in reinforcement learning by \emph{behavioural equivalence}, looking at transitions and rewards rather than raw state features. The concept originated in model checking and MDP minimisation~\citep{givan2003equivalence}: two states are aggregated if they yield the same immediate reward and the same distribution over next equivalent states. Approximate bisimulation metrics~\citep{ferns2004metrics,castro2020scalable} relax the exact equivalence into a distance, extending the idea to high-dimensional environments.

\begin{definition}[$\pi$-bisimulation metric~\citep{castro2020scalable}]\label{def:pi-bisim}
Given two states $s_i,s_j\in\mathcal S$, the $\pi$-bisimulation metric $d_\pi$ measures their behavioural similarity as the fixed point of
\begin{equation}
d_\pi(s_i,s_j)=\underbrace{\big|R^\pi_{s_i}-R^\pi_{s_j}\big|}_{(A)\ \text{reward sufficiency}}\;+\;\underbrace{\gamma\,W_1(d_\pi)\big(P^\pi_{s_i},P^\pi_{s_j}\big)}_{(B)\ \text{transition sufficiency}},
\label{eq:pi-bisim}
\end{equation}
where $R^\pi$ is the reward under policy $\pi$, $P^\pi$ the transition distribution, $\gamma$ the discount factor, and $W_1$ the Wasserstein distance.
\end{definition}

A useful tool for a policy that generalises is knowing which states lead to similar behaviour and which do not; ideally the similarity looks beyond the immediately chosen action to long-term behaviour, which the $\pi$-bisimulation metric does through the full sequence of future rewards. Grounding the metric in rewards, however, can be too restrictive (the policies agree but the obtained rewards do not) or too permissive (the policies differ but the rewards do not)~\citep{agarwal2021psm}. In imitation learning, moreover, no reward is available at all; we instead assume behaviour policies $\pi^\star$ are optimal.

The policy similarity metric (PSM)~\citep{agarwal2021psm} therefore replaces the reward difference in~\Eqref{eq:pi-bisim} with a probability pseudometric $\mathrm{DIST}$ between the policies themselves, keeping the same recursive second term: states are close when the policies at these states, and at their future states, are similar (\citet{agarwal2021psm} ground the metric in the optimal policy). For a given $\mathrm{DIST}$, the PSM $d^*$ satisfies the recursive equation
\begin{equation}
d^*(s_i,s_j)=\underbrace{\mathrm{DIST}\big(\pi(s_i),\pi(s_j)\big)}_{(A)\ \text{policy sufficiency}}\;+\;\underbrace{\gamma\,W_1(d^*)\big(P^\pi_{s_i},P^\pi_{s_j}\big)}_{(B)\ \text{transition sufficiency}},
\label{eq:psm}
\end{equation}
where the discount factor sets the weight between the two terms, and $d^*$ is finite whenever $\mathrm{DIST}$ is bounded. The policy-sufficiency term measures the instantaneous behaviour; the transition-sufficiency term measures the long-term effect carried through the transition dynamics.

\subsection{Memory} \label{sec:optimal_memory}
Instead of the belief state, we consider another representation of the history for visuomotor policies.

\paragraph{Memory is a policy-sufficient statistic.}
We incorporate memory $m_t=\mathcal M(h_t)$ into the graphical model of the POMDP for memory-augmented policies (Figure~\ref{fig:pomdp}(d)),  and seek $\mathcal M$ such that $(o_t,m_t)$ d-separates the expert action $a_t$ from the remaining  history $h_t$, equivalently $I(a_t;h_t\mid o_t,m_t)=0$.
The memory is written deterministically from the history, $p(m_t\mid h_t)=\delta_{\mathcal M(h_t)}(m_t)$; unlike a belief, it is not required to predict the next observation, $p(o_{t+1}\mid m_t,a_t)$, only to render $a_t$ independent of $h_t$ given $(o_t,m_t)$.

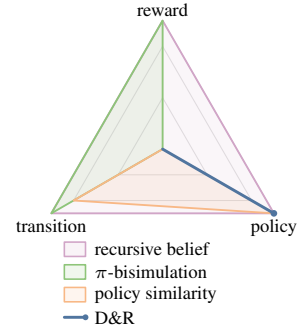
\begin{wrapfigure}{r}{0.3\textwidth}
\vspace{-\intextsep}
\centering
\colorlet{sufBelief}{palLightPurple}
\colorlet{sufBisim}{palLightGreen}
\colorlet{sufPsim}{palLightOrange}
\colorlet{sufOurs}{palBlue}
\begin{tikzpicture}[font=\scriptsize, line cap=round, line join=round]
  \def\R{1.7}
  \foreach \lv in {2,4}{
    \draw[black!12, line width=0.2pt]
      (90:{\R*\lv/5}) -- (210:{\R*\lv/5}) -- (330:{\R*\lv/5}) -- cycle;}
  \foreach \a in {90,210,330}{\draw[black!18, line width=0.2pt] (0,0) -- (\a:\R);}
  \draw[black!25, line width=0.3pt] (90:\R) -- (210:\R) -- (330:\R) -- cycle;
  \fill[sufBelief,opacity=.10] (90:\R) -- (210:\R) -- (330:\R) -- cycle;
  \draw[sufBelief,line width=.7pt] (90:\R) -- (210:\R) -- (330:\R) -- cycle;
  \fill[sufBisim,opacity=.12] (90:\R) -- (210:\R) -- (0,0) -- cycle;
  \draw[sufBisim,line width=.7pt] (90:\R) -- (210:\R) -- (0,0) -- cycle;
  \fill[sufPsim,opacity=.14] (0,0) -- (210:{\R*4/5}) -- (330:\R) -- cycle;
  \draw[sufPsim,line width=.7pt] (0,0) -- (210:{\R*4/5}) -- (330:\R) -- cycle;
  \draw[sufOurs,line width=1.1pt] (0,0) -- (330:\R);
  \fill[sufOurs] (330:\R) circle (1.25pt);
  \node[anchor=south,inner sep=1.5pt] at (90:\R) {reward};
  \node[anchor=north,inner sep=2pt] at (210:\R) {transition};
  \node[anchor=north,inner sep=2pt] at (330:\R) {policy};
  \begin{scope}[shift={(-1.30,-1.42)}]
    \foreach \i/\c in {0/sufBelief, 1/sufBisim, 2/sufPsim}{
      \fill[\c,opacity=.13] (0,{-\i*0.30}) rectangle ++(0.28,0.19);
      \draw[\c,line width=.6pt] (0,{-\i*0.30}) rectangle ++(0.28,0.19);}
    \node[anchor=west,inner sep=1pt] at (0.36,0.095)  {recursive belief};
    \node[anchor=west,inner sep=1pt] at (0.36,-0.205) {$\pi$-bisimulation};
    \node[anchor=west,inner sep=1pt] at (0.36,-0.505) {policy similarity};
    \draw[sufOurs,line width=1.1pt] (0,-0.805) -- (0.28,-0.805);
    \fill[sufOurs] (0.28,-0.805) circle (1.1pt);
    \node[anchor=west,inner sep=1pt] at (0.36,-0.805) {\ours};
  \end{scope}
\end{tikzpicture}
\caption{Different methods assume different sufficiency metrics on the history. Bisimulation measures reward and transition; policy similarity measures policy and transition; the belief takes all three. \ours{} keeps policy sufficiency alone, under a limited budget.}
\label{fig:sufficiency}
\end{wrapfigure}
We measure what information the memory should carry:
\begin{proposition*}[Proposition~\ref{prop:optmem}, restated]
The information that the history $h_t$ carries for control beyond the current observation is the conditional mutual information $I\big(a_t;h_t\mid o_t\big) = \mathbb E_{p(h_t,o_t)}[\KL\big(\pi^\star(\cdot\mid o_t,h_t)\,\big\|\,p(\cdot\mid o_t)\big)]$, the expected divergence between the expert policy and the action distribution obtained by marginalising it over the history. Let the memory be any function of the full history, $m_t=\mathcal M(h_t)$, inserted into the graphical model of Figure~\ref{fig:pomdp}(d). By the chain rule, the memory splits this information into the part it keeps and the part it discards,
\begin{equation}
I\big(a_t;\,h_t\mid o_t\big)
\;=\; I\big(a_t;\,m_t\mid o_t\big)
\;+\;I\big(a_t;\,h_t\mid o_t,m_t\big).
\notag
\end{equation}
The left-hand side of~\Eqref{eq:mem-decomp} is fixed by the task, so the \textbf{optimal memory for visuomotor policies} needs to keep the most action-relevant information and discard the least,
\begin{equation}
\arg\max_{\mathcal M}\; I\big(a_t;\,m_t\mid o_t\big).
\notag
\end{equation}
A memory that maximises \Eqref{eq:keep-equals-lose} is a policy-sufficient statistic (Figure~\ref{fig:sufficiency}), i.e., $\pi^\star(a_t\mid o_t,h_t)=\pi^\star(a_t\mid o_t,m_t)$ for all $(o_t,h_t)$.
\end{proposition*}

The belief is a sufficient statistic of the history: every reward function and every future observation can be predicted from it. A memory need only be sufficient for the policy.

To see what can actually be optimised, we unfold the memory term of Proposition~\ref{prop:optmem}:
\begin{align}
I\big(a_t;\,m_t\mid o_t\big)
& = H(a_t\mid o_t)\;-\;H\big(a_t\mid o_t,m_t\big)
&& \text{\small(definition)}\notag\\[2pt]
&= \mathbb E_{\mathcal D}\big[\log p(a_t\mid o_t,m_t)\big] + H(a_t\mid o_t)
&& \text{\small(expand the residual)}\notag\\[2pt]
&\;\ge\; \mathbb E_{\mathcal D}\big[\log \pi_\theta(a_t\mid o_t,m_t)\big] + H(a_t\mid o_t)
&& \text{\small(variational $\pi_\theta$ for $p$)}\notag
\end{align}
The first two lines are written in terms of the true $p(a_t\mid o_t,m_t)$ of the demonstration distribution and cannot be evaluated directly; replacing it by the policy $\pi_\theta$ gives a variational lower bound, and $H(a_t\mid o_t)$ is a constant of the task. Maximising the memory term is therefore lower-bounded by minimising the action loss,
\begin{equation}
\max_{\mathcal M}\; I\big(a_t;\,m_t\mid o_t\big)
\;\;\ge\;\;
\max_{\mathcal M,\,\theta}\; \mathbb E_{\mathcal D}\big[\log \pi_\theta(a_t\mid o_t,m_t)\big]+H(a_t\mid o_t),\notag
\end{equation}
Policy and memory can thus be trained together end-to-end against a single action loss, with no auxiliary objective and no supervision on what to remember.

\paragraph{Memory under the policy similarity metric.}
We revisit the policy similarity metric of Equation~\ref{eq:psm}, taking $\mathrm{DIST}$ to be the total variation distance. A memory is a compressed history, so we measure it against the history it compresses:
\begin{equation}
d^*(m_t,h_t)=\underbrace{\mathrm{TV}\big(\pi^\star(\cdot\mid o_t,m_t),\ \pi^\star(\cdot\mid o_t,h_t)\big)}_{(A)\ \text{policy sufficiency}}\;+\;\underbrace{W_1(d^*)\big(P^{\pi^\star}_{o_t,m_t},\,P^{\pi^\star}_{h_t}\big)}_{(B)\ \text{transition sufficiency}},
\label{eq:memory-psm}
\end{equation}
where we drop $\gamma$: in RL it is given by the discount factor, and LfD has none. Term (A) is related to the quantity Proposition~\ref{prop:optmem} optimises, and term (B) is left out.

\clearpage

\clearpage
\section{Experimental details}\label{app:exp}

\subsection{Experimental setup}

We follow a unified training recipe across all methods to ensure a fair comparison. Specifically, we adopt the released $\pi_{0.5}$ configuration from the RoboMME benchmark and exclude proprioceptive states from the inputs, as they bring no benefit. To accelerate training, we freeze the SigLIP vision backbone and precompute and cache all visual tokens, fine-tuning only the VLM expert, the action expert, and any parameters introduced by the memory (for \ours, the selector). All models are trained for 80K steps; the perceptual and latent memory variants use a batch size of 64, while the parametric (TTT) variants use 16. All models are trained on $4\times$A800 GPUs with cached data. Training takes about $20$ hours for the perceptual and latent memory variants, $45$ hours for TTT, and $10$ days for RB-VLA. Detailed hyperparameters are provided in Table~\ref{tab:hyperparams}.

\begin{table}[H]
\centering
\footnotesize
\caption{Training hyperparameters.}
\label{tab:hyperparams}
\begin{tabular}{@{}ll@{}}
\toprule
\textbf{Hyperparameter} & \textbf{Value} \\
\midrule
Base model & $\pi_{0.5}$ base \\
Batch size & $64$ global \\
Action horizon & 20 \\
Action space & Joint-space \\
Proprioception & False \\
Total steps & 80,000 \\
Warmup steps & 10,000 \\
LR & $5\times10^{-5}$ \\
LR schedule & Constant \\
Optimiser & AdamW ($\beta_1{=}0.9$, $\beta_2{=}0.95$, weight decay 0) \\
Gradient clip norm & 1.0 \\
EMA decay & 0.999 \\
Frozen part & SigLIP ViT \\
Hardware & $4\times$ NVIDIA A800, $80$\,GB \\
Parallelism & FSDP across the $4$ GPUs \\
\bottomrule
\end{tabular}
\end{table}

\subsection{Variants of \ours}\label{app:variants}

\paragraph{\ours{}+pred.} This variant adds RB-VLA's world-model objective~\citep{bagaria2026rbvla} to \ours, leaving the selector, the reduction tree and the policy unchanged. For every frame $f$ of the candidate pool we form a frame latent $z_f\in\mathbb{R}^{512}$: the mean of the frame's patch tokens after the memory encoder, RMS-normalised and linearly projected. Two predictors map the two-frame context $[z_{f-1}, z_f]$ and a recency feature of frame $f$ to the next latent and to the latent $h$ pool frames ahead,
\begin{equation}
\mathcal{L}_{\mathrm{wm}}=\mathbb{E}_f\big[\lVert \hat z_{f+1}-\mathrm{sg}(z_{f+1})\rVert^2\big]+\tfrac12\,\mathbb{E}_f\big[\lVert \hat z_{f+h}-\mathrm{sg}(z_{f+h})\rVert^2\big],
\end{equation}
where $\mathrm{sg}$ is the stop-gradient and the expectation runs over valid (non-padded) frame pairs; we use $h=4$ on the $16$-frame pool. The two predictors are two-layer MLPs with SiLU activations, and the pool budget, the selector, and the policy are identical to \ours. The total loss is $\mathcal{L}_{\mathrm{action}}+\mathcal{L}_{\mathrm{wm}}$, and the gradient of $\mathcal{L}_{\mathrm{wm}}$ reaches the memory encoder through the predictors' inputs, so the memory features are asked to be predictive of the future as well as sufficient for the action. The stop-gradient targets and the $1{:}\tfrac12$ weighting of the one-step and $h$-step terms follow RB-VLA; we omit its inverse-dynamics and KL terms, which require per-step actions and a stochastic belief that a static candidate pool does not have.

Table~\ref{tab:dnr-pool} sweeps the size of the candidate pool (Table~\ref{tab:dnr-pool-taskchar} regroups the same runs by task characteristic), and Figure~\ref{fig:analysis}(c) plots the same runs per task characteristic; Table~\ref{tab:dnr-ablations} ablates the training signal and the candidate set at a fixed pool of $|\mathcal{H}|=256$, against \ours at the same pool size.

\begin{table}[t]
\centering\small
\caption{\textbf{Candidate pool.} Success rate (\%) of \ours averaged over the four tasks of each suite, as
mean\,$\pm$\,standard deviation over evaluation runs (checkpoints $\times$ seeds $0/7/42$), as the size $|\mathcal{H}|$ of the
candidate pool the selector reduces grows from $8$ to $64$ frames. Every row uses a memory budget of $K=64$ tokens. \textbf{bold} is the highest success rate in each column.}
\label{tab:dnr-pool}
\setlength{\tabcolsep}{5pt}\renewcommand{\arraystretch}{1.1}
\begin{tabular}{@{}l ccccc@{}}
\toprule
Variant & Counting & Permanence & Reference & Imitation & AVG \\
\midrule
\quad $|\mathcal{H}|=128$\ \ (8 frames) & 58.25\,$\pm$\,2.04 & 20.83\,$\pm$\,4.79 & 27.00\,$\pm$\,1.38 & 34.67\,$\pm$\,1.70 & 35.19\,$\pm$\,1.98 \\
\quad $|\mathcal{H}|=256$\ \ (16 frames) & 59.17\,$\pm$\,2.84 & 25.17\,$\pm$\,2.47 & 24.56\,$\pm$\,2.13 & \textbf{38.67\,$\pm$\,1.99} & 36.89\,$\pm$\,0.81 \\
\quad $|\mathcal{H}|=512$\ \ (32 frames) & 61.17\,$\pm$\,0.62 & \textbf{26.83\,$\pm$\,1.03} & \textbf{28.50\,$\pm$\,1.41} & 37.83\,$\pm$\,2.78 & \textbf{38.58\,$\pm$\,0.50} \\
\quad $|\mathcal{H}|=1024$ (64 frames) & \textbf{61.89\,$\pm$\,2.37} & 26.17\,$\pm$\,2.32 & 20.94\,$\pm$\,1.57 & 28.67\,$\pm$\,2.58 & 34.42\,$\pm$\,1.03 \\
\addlinespace[2pt]
\bottomrule
\end{tabular}
\end{table}

\begin{table}[t]
\centering\small
\caption{\textbf{Candidate pool by task characteristic.} Success rate (\%) of \ours averaged over the tasks of each characteristic of \citet{dai2026robomme} (the grouping of Table~\ref{tab:robomme-taskchar}), as mean\,$\pm$\,standard deviation over evaluation runs (checkpoints $\times$ seeds $0/7/42$), as the size $|\mathcal{H}|$ of the candidate pool changes. Every row uses a memory budget of $K=64$ tokens; the AVG column is that of Table~\ref{tab:dnr-pool}. \textbf{bold} is the highest success rate in each column.}
\label{tab:dnr-pool-taskchar}
\setlength{\tabcolsep}{4pt}\renewcommand{\arraystretch}{1.1}
\resizebox{\linewidth}{!}{%
\begin{tabular}{@{}l ccccccc@{}}
\toprule
Variant & \makecell{Motion-\\Centric} & \makecell{Time-\\Sensitive} & \makecell{Short-Horizon\\Video} & \makecell{Long-Horizon\\Video} & \makecell{Dynamic\\Scene-Change} & \makecell{Event-\\Salient} & AVG \\
\midrule
\quad $|\mathcal{H}|=128$\ \ (8 frames) & 41.92\,$\pm$\,1.62 & 20.67\,$\pm$\,8.46 & 26.33\,$\pm$\,1.58 & \textbf{33.17\,$\pm$\,1.77} & 15.33\,$\pm$\,5.84 & 61.11\,$\pm$\,2.69 & 35.19\,$\pm$\,1.98 \\
\quad $|\mathcal{H}|=256$\ \ (16 frames) & 45.94\,$\pm$\,2.05 & 28.00\,$\pm$\,7.06 & 26.52\,$\pm$\,1.80 & 30.67\,$\pm$\,3.06 & 19.33\,$\pm$\,1.99 & 59.85\,$\pm$\,2.77 & 36.89\,$\pm$\,0.81 \\
\quad $|\mathcal{H}|=512$\ \ (32 frames) & \textbf{47.50\,$\pm$\,1.08} & 31.33\,$\pm$\,2.49 & \textbf{28.89\,$\pm$\,1.37} & 31.67\,$\pm$\,1.70 & \textbf{23.78\,$\pm$\,1.26} & 58.22\,$\pm$\,1.66 & \textbf{38.58\,$\pm$\,0.50} \\
\quad $|\mathcal{H}|=1024$ (64 frames) & 36.17\,$\pm$\,1.41 & \textbf{32.44\,$\pm$\,5.56} & 23.19\,$\pm$\,2.06 & 25.78\,$\pm$\,2.15 & 22.44\,$\pm$\,2.95 & \textbf{61.70\,$\pm$\,3.02} & 34.42\,$\pm$\,1.03 \\
\addlinespace[2pt]
\bottomrule
\end{tabular}}
\end{table}

\begin{table}[t]
\centering\small
\caption{\textbf{Ablations.} Success rate (\%) of \ours and its variants at a candidate pool of $|\mathcal{H}|=256$, as mean\,$\pm$\,standard deviation over evaluation runs (checkpoints $\times$ seeds $0/7/42$), averaged over the four tasks of each suite (top) and per task (bottom). $+$pred adds RB-VLA's world-model loss to the action loss; $+$text adds the language instruction to the candidate pool the selector scores. Every row uses a memory budget of $K=64$ tokens. \textbf{bold} is the highest success rate in each column.}
\label{tab:dnr-ablations}
\setlength{\tabcolsep}{5pt}\renewcommand{\arraystretch}{1.1}
\begin{tabular}{@{}l ccccc@{}}
\toprule
Variant & Counting & Permanence & Reference & Imitation & AVG \\
\midrule
\ours\ \ (reference) & \textbf{59.17\,$\pm$\,2.84} & \textbf{25.17\,$\pm$\,2.47} & \textbf{24.56\,$\pm$\,2.13} & \textbf{38.67\,$\pm$\,1.99} & \textbf{36.89\,$\pm$\,0.81} \\
\addlinespace[2pt]
\quad $+$pred & 58.62\,$\pm$\,1.85 & 22.69\,$\pm$\,1.52 & 22.12\,$\pm$\,1.43 & 30.75\,$\pm$\,2.96 & 33.55\,$\pm$\,0.70 \\
\quad $+$text & 52.00\,$\pm$\,2.31 & 21.94\,$\pm$\,3.51 & 23.11\,$\pm$\,2.02 & 24.22\,$\pm$\,3.15 & 30.32\,$\pm$\,1.54 \\
\addlinespace[2pt]
\bottomrule
\end{tabular}

\vspace{6pt}
\setlength{\tabcolsep}{3pt}
\resizebox{\linewidth}{!}{%
\begin{tabular}{@{}l cccc cccc@{}}
\toprule
 & \multicolumn{4}{c}{\textbf{Counting}} & \multicolumn{4}{c}{\textbf{Permanence}} \\
\cmidrule(lr){2-5}\cmidrule(lr){6-9}
Variant & \makecell{BinFill} & \makecell{PickXtimes} & \makecell{SwingXtimes} & \makecell{StopCube} & \makecell{VideoUnmask} & \makecell{ButtonUnmask} & \makecell{VideoUnmaskSwap} & \makecell{ButtonUnmaskSwap} \\
\midrule
\ours\ \ (reference) & 38.44\,$\pm$\,4.79 & 78.67\,$\pm$\,4.00 & 91.56\,$\pm$\,3.74 & \textbf{28.00\,$\pm$\,7.06} & \textbf{35.11\,$\pm$\,6.74} & 24.22\,$\pm$\,4.57 & \textbf{24.89\,$\pm$\,3.41} & 16.44\,$\pm$\,6.24 \\
\addlinespace[2pt]
\quad $+$pred & \textbf{41.25\,$\pm$\,4.47} & \textbf{80.00\,$\pm$\,5.00} & \textbf{94.75\,$\pm$\,2.44} & 18.50\,$\pm$\,5.81 & 31.00\,$\pm$\,3.74 & \textbf{25.75\,$\pm$\,3.23} & 15.75\,$\pm$\,4.41 & \textbf{18.25\,$\pm$\,2.73} \\
\quad $+$text & 38.44\,$\pm$\,2.95 & 68.89\,$\pm$\,8.60 & 83.78\,$\pm$\,6.76 & 16.89\,$\pm$\,4.01 & 34.00\,$\pm$\,3.27 & 19.78\,$\pm$\,5.77 & 22.67\,$\pm$\,3.13 & 11.33\,$\pm$\,6.32 \\
\bottomrule
\end{tabular}}

\resizebox{\linewidth}{!}{%
\begin{tabular}{@{}l cccc cccc@{}}
\toprule
 & \multicolumn{4}{c}{\textbf{Reference}} & \multicolumn{4}{c}{\textbf{Imitation}} \\
\cmidrule(lr){2-5}\cmidrule(lr){6-9}
Variant & \makecell{PickHighlight} & \makecell{VideoRepick} & \makecell{VideoPlaceButton} & \makecell{VideoPlaceOrder} & \makecell{MoveCube} & \makecell{InsertPeg} & \makecell{PatternLock} & \makecell{RouteStick} \\
\midrule
\ours\ \ (reference) & \textbf{17.33\,$\pm$\,4.32} & 19.56\,$\pm$\,3.24 & \textbf{39.78\,$\pm$\,5.12} & 21.56\,$\pm$\,4.30 & \textbf{62.44\,$\pm$\,6.02} & \textbf{4.67\,$\pm$\,1.33} & \textbf{32.00\,$\pm$\,7.72} & \textbf{55.56\,$\pm$\,3.74} \\
\addlinespace[2pt]
\quad $+$pred & 15.50\,$\pm$\,4.21 & 19.25\,$\pm$\,3.73 & 27.00\,$\pm$\,3.16 & \textbf{26.75\,$\pm$\,2.82} & 50.25\,$\pm$\,2.73 & 1.75\,$\pm$\,1.85 & 27.50\,$\pm$\,7.33 & 43.50\,$\pm$\,3.97 \\
\quad $+$text & 16.22\,$\pm$\,4.57 & \textbf{21.11\,$\pm$\,4.01} & 32.22\,$\pm$\,3.94 & 22.89\,$\pm$\,5.74 & 50.89\,$\pm$\,6.33 & 1.11\,$\pm$\,0.99 & 20.22\,$\pm$\,4.66 & 24.67\,$\pm$\,4.62 \\
\bottomrule
\end{tabular}}

\end{table}

\subsection{Additional simulation results}\label{app:sim-results}

\subsubsection{Full results on all tasks}\label{app:full-results}

We report results by suite: counting, permanence, reference and imitation, which ask the policy to accumulate a count over repeated events, to track an object after it is hidden, to refer back to an earlier moment of the episode, and to reproduce a demonstrated sequence.
Table~\ref{tab:full-results} reports every method by suite and by task, with standard deviations over seeds; the main text groups the same numbers by task characteristic (Table~\ref{tab:robomme-taskchar}).

\begin{table}[H]
\centering\footnotesize
\caption{\textbf{Full results on RoboMME.} Success rate (\%) of every method under the $64$-token budget, grouped by what the memory stores. Top: mean over the four tasks of each suite and over all $16$ tasks (AVG). Bottom: every task, as mean $\pm$ standard deviation over seeds. The $\pi_{0.5}$ row is the number reported by \citet{dai2026robomme}. \textbf{bold} is the highest success rate in each column.}
\label{tab:full-results}
\setlength{\tabcolsep}{3pt}\renewcommand{\arraystretch}{1.12}
{\setlength{\tabcolsep}{5pt}
\begin{tabular}{@{}l l ccccc@{}}
\toprule
Type & Method & \textbf{Counting} & \textbf{Permanence} & \textbf{Reference} & \textbf{Imitation} & \textbf{AVG} \\
\midrule
No memory & $\pi_{0.5}$ & 28.78 & 17.00 & 17.17 & 8.78 & 17.93 \\
\midrule
\multirow{2}{*}{Perceptual} & FrameSamp & 52.67 & 15.67 & 21.34 & 22.06 & 27.93 \\
 & TokenDrop & 27.50 & 17.11 & 15.45 & 11.33 & 17.85 \\
\midrule
\multirow{2}{*}{Latent} & HAMLET & 60.00 & 25.34 & 24.06 & 19.50 & 32.22 \\
 & RB-VLA & 53.17 & 25.33 & 25.34 & 10.22 & 28.52 \\
\midrule
Parametric & TTT & 28.84 & 16.72 & 13.61 & 9.34 & 17.13 \\
\midrule
Ours & \ours & \textbf{61.17} & \textbf{26.83} & \textbf{28.50} & \textbf{37.83} & \textbf{38.58} \\
\bottomrule
\end{tabular}}

\vspace{6pt}
\resizebox{\linewidth}{!}{%
\begin{tabular}{@{}l l cccc cccc@{}}
\toprule
 & & \multicolumn{4}{c}{\textbf{Counting}} & \multicolumn{4}{c}{\textbf{Permanence}} \\
\cmidrule(lr){3-6}\cmidrule(lr){7-10}
Type & Method & \makecell{BinFill} & \makecell{PickXtimes} & \makecell{SwingXtimes} & \makecell{StopCube} & \makecell{VideoUnmask} & \makecell{ButtonUnmask} & \makecell{VideoUnmaskSwap} & \makecell{ButtonUnmaskSwap} \\
\midrule
No memory & $\pi_{0.5}$ & 30.00\,$\pm$\,6.48 & 42.89\,$\pm$\,6.41 & 35.56\,$\pm$\,5.27 & 6.67\,$\pm$\,5.20 & 20.44\,$\pm$\,2.96 & 22.22\,$\pm$\,10.22 & 18.67\,$\pm$\,4.24 & 6.67\,$\pm$\,4.12 \\
\midrule
\multirow{2}{*}{Perceptual} & FrameSamp & 34.44\,$\pm$\,2.57 & 60.44\,$\pm$\,3.10 & 88.45\,$\pm$\,3.00 & 27.34\,$\pm$\,7.36 & 30.00\,$\pm$\,3.31 & 6.67\,$\pm$\,2.18 & 20.89\,$\pm$\,1.57 & 5.11\,$\pm$\,2.20 \\
 & TokenDrop & 27.78\,$\pm$\,4.29 & 39.78\,$\pm$\,6.82 & 36.67\,$\pm$\,5.57 & 5.78\,$\pm$\,5.33 & 28.67\,$\pm$\,2.83 & 16.67\,$\pm$\,6.48 & 16.67\,$\pm$\,3.61 & 6.44\,$\pm$\,4.67 \\
\midrule
\multirow{2}{*}{Latent} & HAMLET & 34.89\,$\pm$\,3.48 & \textbf{84.67\,$\pm$\,6.86} & 81.11\,$\pm$\,6.86 & \textbf{39.33\,$\pm$\,6.78} & 27.78\,$\pm$\,2.73 & 26.67\,$\pm$\,3.32 & 25.11\,$\pm$\,1.76 & \textbf{21.78\,$\pm$\,6.28} \\
 & RB-VLA & \textbf{36.67\,$\pm$\,5.03} & 79.33\,$\pm$\,2.31 & 86.00\,$\pm$\,5.29 & 10.67\,$\pm$\,4.62 & 30.00\,$\pm$\,3.46 & \textbf{28.00\,$\pm$\,3.46} & 23.33\,$\pm$\,1.15 & 20.00\,$\pm$\,4.00 \\
\midrule
Parametric & TTT & 28.22\,$\pm$\,6.82 & 42.67\,$\pm$\,4.80 & 41.56\,$\pm$\,5.46 & 2.89\,$\pm$\,2.67 & 28.67\,$\pm$\,4.69 & 15.11\,$\pm$\,8.31 & 19.78\,$\pm$\,2.33 & 3.33\,$\pm$\,4.36 \\
\midrule
Ours & \ours & 35.33\,$\pm$\,0.94 & 82.67\,$\pm$\,1.89 & \textbf{95.33\,$\pm$\,3.77} & 31.33\,$\pm$\,2.49 & \textbf{32.67\,$\pm$\,1.89} & 26.00\,$\pm$\,0.30 & \textbf{27.33\,$\pm$\,3.40} & 21.33\,$\pm$\,4.11 \\
\bottomrule
\end{tabular}}

\resizebox{\linewidth}{!}{%
\begin{tabular}{@{}l l cccc cccc@{}}
\toprule
 & & \multicolumn{4}{c}{\textbf{Reference}} & \multicolumn{4}{c}{\textbf{Imitation}} \\
\cmidrule(lr){3-6}\cmidrule(lr){7-10}
Type & Method & \makecell{PickHighlight} & \makecell{VideoRepick} & \makecell{VideoPlaceButton} & \makecell{VideoPlaceOrder} & \makecell{MoveCube} & \makecell{InsertPeg} & \makecell{PatternLock} & \makecell{RouteStick} \\
\midrule
No memory & $\pi_{0.5}$ & 11.33\,$\pm$\,4.90 & 0.44\,$\pm$\,0.88 & 31.11\,$\pm$\,4.01 & 25.78\,$\pm$\,1.86 & 26.00\,$\pm$\,3.61 & 1.56\,$\pm$\,1.94 & 2.89\,$\pm$\,2.26 & 4.67\,$\pm$\,2.45 \\
\midrule
\multirow{2}{*}{Perceptual} & FrameSamp & 18.67\,$\pm$\,2.37 & 14.22\,$\pm$\,1.13 & 26.89\,$\pm$\,1.66 & 25.56\,$\pm$\,3.86 & 52.66\,$\pm$\,0.94 & 2.22\,$\pm$\,1.37 & 15.56\,$\pm$\,0.83 & 17.78\,$\pm$\,1.13 \\
 & TokenDrop & 14.00\,$\pm$\,3.74 & 1.56\,$\pm$\,1.33 & 26.67\,$\pm$\,3.00 & 19.56\,$\pm$\,4.33 & 24.44\,$\pm$\,4.56 & 2.44\,$\pm$\,2.19 & 9.56\,$\pm$\,2.96 & 8.89\,$\pm$\,3.02 \\
\midrule
\multirow{2}{*}{Latent} & HAMLET & 26.00\,$\pm$\,4.24 & 20.67\,$\pm$\,2.65 & 27.78\,$\pm$\,5.33 & 21.78\,$\pm$\,2.73 & \textbf{60.00\,$\pm$\,4.80} & 1.11\,$\pm$\,1.45 & 7.33\,$\pm$\,2.24 & 9.56\,$\pm$\,2.96 \\
 & RB-VLA & \textbf{26.67\,$\pm$\,9.45} & 26.00\,$\pm$\,2.00 & 24.67\,$\pm$\,4.16 & 24.00\,$\pm$\,2.00 & 36.67\,$\pm$\,1.15 & \textbf{3.33\,$\pm$\,1.15} & 0.67\,$\pm$\,1.15 & 0.20\,$\pm$\,0.01 \\
\midrule
Parametric & TTT & 12.44\,$\pm$\,3.13 & 4.44\,$\pm$\,2.79 & 20.44\,$\pm$\,4.45 & 17.11\,$\pm$\,9.49 & 24.67\,$\pm$\,4.58 & 0.89\,$\pm$\,1.45 & 4.67\,$\pm$\,2.24 & 7.11\,$\pm$\,2.26 \\
\midrule
Ours & \ours & 24.00\,$\pm$\,1.63 & \textbf{26.67\,$\pm$\,0.94} & \textbf{34.00\,$\pm$\,0.40} & \textbf{29.33\,$\pm$\,3.40} & 56.67\,$\pm$\,4.71 & \textbf{3.33\,$\pm$\,0.94} & \textbf{32.00\,$\pm$\,5.89} & \textbf{59.33\,$\pm$\,3.40} \\
\bottomrule
\end{tabular}}

\end{table}

\subsubsection{Ablation: memory budget} \label{app:memory_budget}

Table~\ref{tab:memory_budget_ablation} reports the average success rate as the memory budget grows from $64$ to $512$ tokens, overall and per suite.

\begin{table}[H]
\centering\footnotesize
\caption{\textbf{Effect of the memory budget.} Average success rate (\%) on RoboMME as the memory budget grows from $64$ to $512$ tokens, overall and per suite. The FrameSamp entries from $128$ to $512$ tokens and the TokenDrop entries at $128$ and $256$ tokens are the numbers reported by \citet{dai2026robomme}; the remaining entries are our runs.}
\label{tab:memory_budget_ablation}
\setlength{\tabcolsep}{3pt}\renewcommand{\arraystretch}{1.1}
\begin{tabular}{@{}c cccc@{}}
\toprule
\multicolumn{5}{c}{\textbf{Overall}} \\
\cmidrule(lr){1-5}
Budget & FrameSamp & TokenDrop & HAMLET & \ours \\
\midrule
64 & 27.93 & 17.85 & 32.22 & 38.58 \\
128 & 36.54 & 32.76 & 33.00 & 49.01 \\ 
256 & 42.90 & 36.36 & 32.50 & 50.11 \\
512 & 44.51 & 36.27 & 36.06 & 46.40 \\ 
\bottomrule
\end{tabular}
\end{table}

Table~\ref{tab:robomme-pertask-512} repeats the per-task comparison of Table~\ref{tab:full-results} with a memory budget of $512$ tokens.

\begin{table}[H]
\centering\footnotesize
\caption{\textbf{Results with a memory budget of $512$ tokens.} Per-task success rate (\%) on RoboMME~\citep{dai2026robomme} as mean $\pm$ standard deviation over seeds, with $\pi_{0.5}$ as the backbone and methods grouped by what the memory stores. The $\pi_{0.5}$, FrameSamp, and TTT rows are the numbers reported by \citet{dai2026robomme}. \textbf{bold} is the highest success rate in each column. }
\label{tab:robomme-pertask-512}
\setlength{\tabcolsep}{3pt}\renewcommand{\arraystretch}{1.12}
\resizebox{\linewidth}{!}{%
\begin{tabular}{@{}l l cccc cccc@{}}
\toprule
 & & \multicolumn{4}{c}{\textbf{Counting}} & \multicolumn{4}{c}{\textbf{Permanence}} \\
\cmidrule(lr){3-6}\cmidrule(lr){7-10}
Type & Method & \makecell{BinFill} & \makecell{PickXtimes} & \makecell{SwingXtimes} & \makecell{StopCube} & \makecell{VideoUnmask} & \makecell{ButtonUnmask} & \makecell{VideoUnmaskSwap} & \makecell{ButtonUnmaskSwap} \\
\midrule
No memory & $\pi_{0.5}$ & 30.00\,$\pm$\,6.48 & 42.89\,$\pm$\,6.41 & 35.56\,$\pm$\,5.27 & 6.67\,$\pm$\,5.20 & 20.44\,$\pm$\,2.96 & 22.22\,$\pm$\,10.22 & 18.67\,$\pm$\,4.24 & 6.67\,$\pm$\,4.12 \\
\midrule
\multirow{2}{*}{Perceptual} & FrameSamp & 39.56\,$\pm$\,5.27 & 87.33\,$\pm$\,2.45 & \textbf{92.00\,$\pm$\,2.24} & 42.00\,$\pm$\,11.36 & \textbf{32.67\,$\pm$\,3.16} & 25.11\,$\pm$\,3.18 & 24.44\,$\pm$\,6.06 & 18.22\,$\pm$\,3.80 \\
 & TokenDrop & 31.33\,$\pm$\,6.02 & 84.67\,$\pm$\,4.84 & 86.00\,$\pm$\,4.56 & 11.00\,$\pm$\,4.69 & 30.67\,$\pm$\,2.07 & 23.67\,$\pm$\,4.46 & 22.67\,$\pm$\,5.01 & 21.00\,$\pm$\,6.03 \\
\midrule
\multirow{2}{*}{Latent} & HAMLET & 31.78\,$\pm$\,2.91 & \textbf{88.89\,$\pm$\,3.48} & 89.33\,$\pm$\,3.74 & \textbf{52.67\,$\pm$\,7.42} & 29.78\,$\pm$\,3.23 & 25.11\,$\pm$\,4.26 & 23.33\,$\pm$\,3.32 & 22.67\,$\pm$\,3.61 \\
 & RB-VLA & \textbf{39.67\,$\pm$\,3.13} & 78.31\,$\pm$\,5.01 & 81.00\,$\pm$\,8.50 & 12.02\,$\pm$\,1.15 & 30.01\,$\pm$\,3.21 & 24.20\,$\pm$\,2.66 & 23.65\,$\pm$\,0.13 & 19.10\,$\pm$\,3.00 \\
\midrule
\multirow{1}{*}{Parametric} & TTT & 34.89\,$\pm$\,6.57 & 63.78\,$\pm$\,6.12 & 41.33\,$\pm$\,9.95 & 4.00\,$\pm$\,3.46 & 31.78\,$\pm$\,6.12 & 22.44\,$\pm$\,3.13 & 19.56\,$\pm$\,6.98 & 18.00\,$\pm$\,6.86 \\
\midrule
\multirow{2}{*}{Ours} & \textbf{\ours} & 38.00\,$\pm$\,1.63 & 68.67\,$\pm$\,0.94 & \textbf{92.00\,$\pm$\,4.32} & 48.67\,$\pm$\,0.94 & 26.67\,$\pm$\,2.49 & \textbf{26.00\,$\pm$\,1.63} & \textbf{30.67\,$\pm$\,1.89} & \textbf{30.00\,$\pm$\,5.89} \\
\bottomrule
\end{tabular}}

\vspace{6pt}
\resizebox{\linewidth}{!}{%
\begin{tabular}{@{}l l cccc cccc@{}}
\toprule
 & & \multicolumn{4}{c}{\textbf{Reference}} & \multicolumn{4}{c}{\textbf{Imitation}} \\
\cmidrule(lr){3-6}\cmidrule(lr){7-10}
Type & Method & \makecell{PickHighlight} & \makecell{VideoRepick} & \makecell{VideoPlaceButton} & \makecell{VideoPlaceOrder} & \makecell{MoveCube} & \makecell{InsertPeg} & \makecell{PatternLock} & \makecell{RouteStick} \\
\midrule
No memory & $\pi_{0.5}$ & 11.33\,$\pm$\,4.90 & 0.44\,$\pm$\,0.88 & 31.11\,$\pm$\,4.01 & 25.78\,$\pm$\,1.86 & 26.00\,$\pm$\,3.61 & 1.56\,$\pm$\,1.94 & 2.89\,$\pm$\,2.26 & 4.67\,$\pm$\,2.45 \\
\midrule
\multirow{2}{*}{Perceptual} & FrameSamp & 22.89\,$\pm$\,3.89 & 30.44\,$\pm$\,5.81 & \textbf{60.00\,$\pm$\,4.00} & 32.00\,$\pm$\,3.87 & 77.78\,$\pm$\,3.80 & 7.56\,$\pm$\,3.57 & \textbf{53.56\,$\pm$\,4.56} & \textbf{66.67\,$\pm$\,4.12} \\
 & TokenDrop & 21.00\,$\pm$\,5.02 & 25.67\,$\pm$\,4.63 & 45.67\,$\pm$\,4.27 & 30.33\,$\pm$\,4.63 & 70.33\,$\pm$\,5.85 & 3.67\,$\pm$\,2.94 & 21.67\,$\pm$\,3.44 & 51.00\,$\pm$\,6.29 \\
\midrule
\multirow{2}{*}{Latent} & HAMLET & 24.67\,$\pm$\,6.00 & 25.11\,$\pm$\,2.85 & 34.00\,$\pm$\,2.45 & 21.56\,$\pm$\,2.96 & 75.33\,$\pm$\,7.00 & 6.22\,$\pm$\,3.07 & 8.00\,$\pm$\,3.46 & 18.22\,$\pm$\,2.33 \\
 & RB-VLA & 24.37\,$\pm$\,9.35 & 23.01\,$\pm$\,4.01 & 23.77\,$\pm$\,5.01 & 23.01\,$\pm$\,2.10 & 38.61\,$\pm$\,1.10 & 2.31\,$\pm$\,1.05 & 0.61\,$\pm$\,1.30 & 0.01\,$\pm$\,0.30 \\
\midrule
\multirow{1}{*}{Parametric} & TTT & 12.22\,$\pm$\,3.80 & 9.56\,$\pm$\,3.43 & 34.00\,$\pm$\,6.00 & 22.89\,$\pm$\,4.59 & 33.56\,$\pm$\,6.06 & 0.89\,$\pm$\,1.05 & 3.11\,$\pm$\,2.85 & 5.56\,$\pm$\,3.28 \\
\midrule
\multirow{2}{*}{Ours} & \textbf{\ours} & \textbf{33.33\,$\pm$\,1.89} & \textbf{52.00\,$\pm$\,5.89} & 54.00\,$\pm$\,1.63 & \textbf{34.67\,$\pm$\,2.49} & \textbf{86.00\,$\pm$\,2.83} & \textbf{7.67\,$\pm$\,0.47} & 53.33\,$\pm$\,2.49 & 60.67\,$\pm$\,3.40 \\
\bottomrule
\end{tabular}}
\end{table}

\subsubsection{Memory visualisation on all tasks}\label{app:memory_vis}

\begin{figure}[H]
\centering
\vspace{-2pt}
\includegraphics[width=0.78\linewidth]{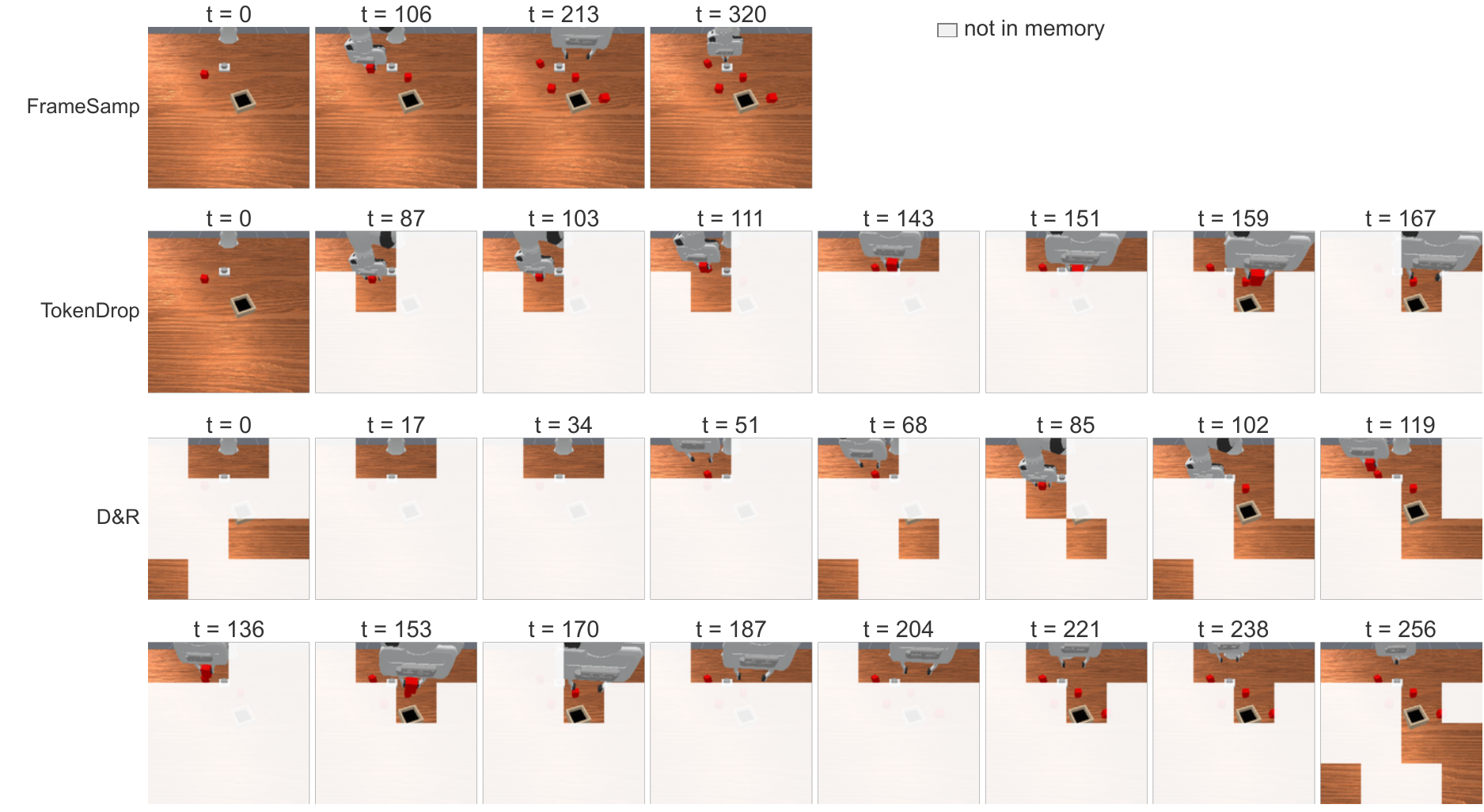}
\vspace{-4pt}
\caption{\textbf{BinFill} (Counting): ``put one red cube into the bin, then press the button to stop''.}
\label{fig:memory_vis_BinFill}
\end{figure}

\begin{figure}[H]
\centering
\vspace{-2pt}
\includegraphics[width=0.78\linewidth]{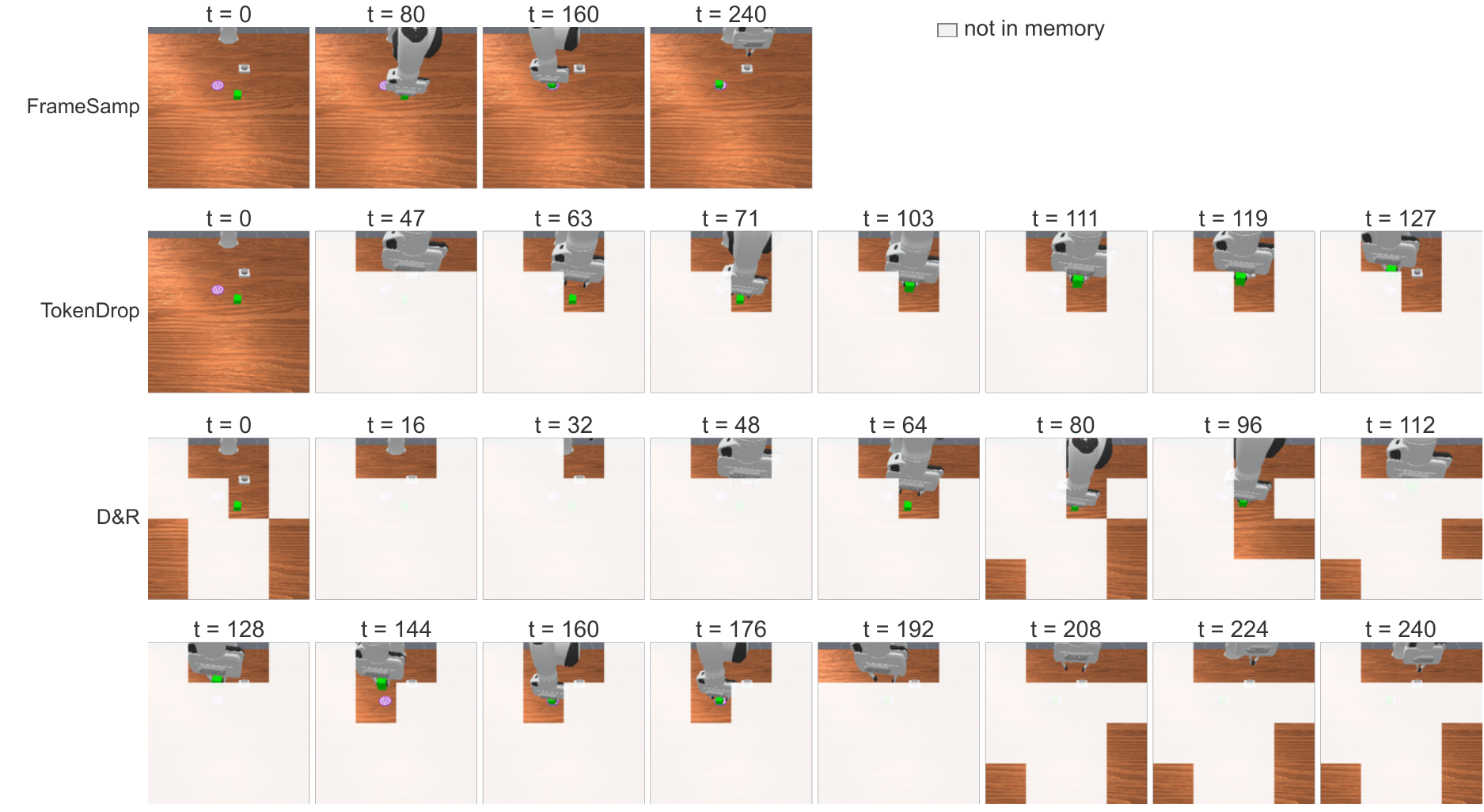}
\vspace{-4pt}
\caption{\textbf{PickXtimes} (Counting): ``pick up the green cube and place it on the target, then press the button to stop''.}
\label{fig:memory_vis_PickXtimes}
\end{figure}

\begin{figure}[H]
\centering
\vspace{-2pt}
\includegraphics[width=0.78\linewidth]{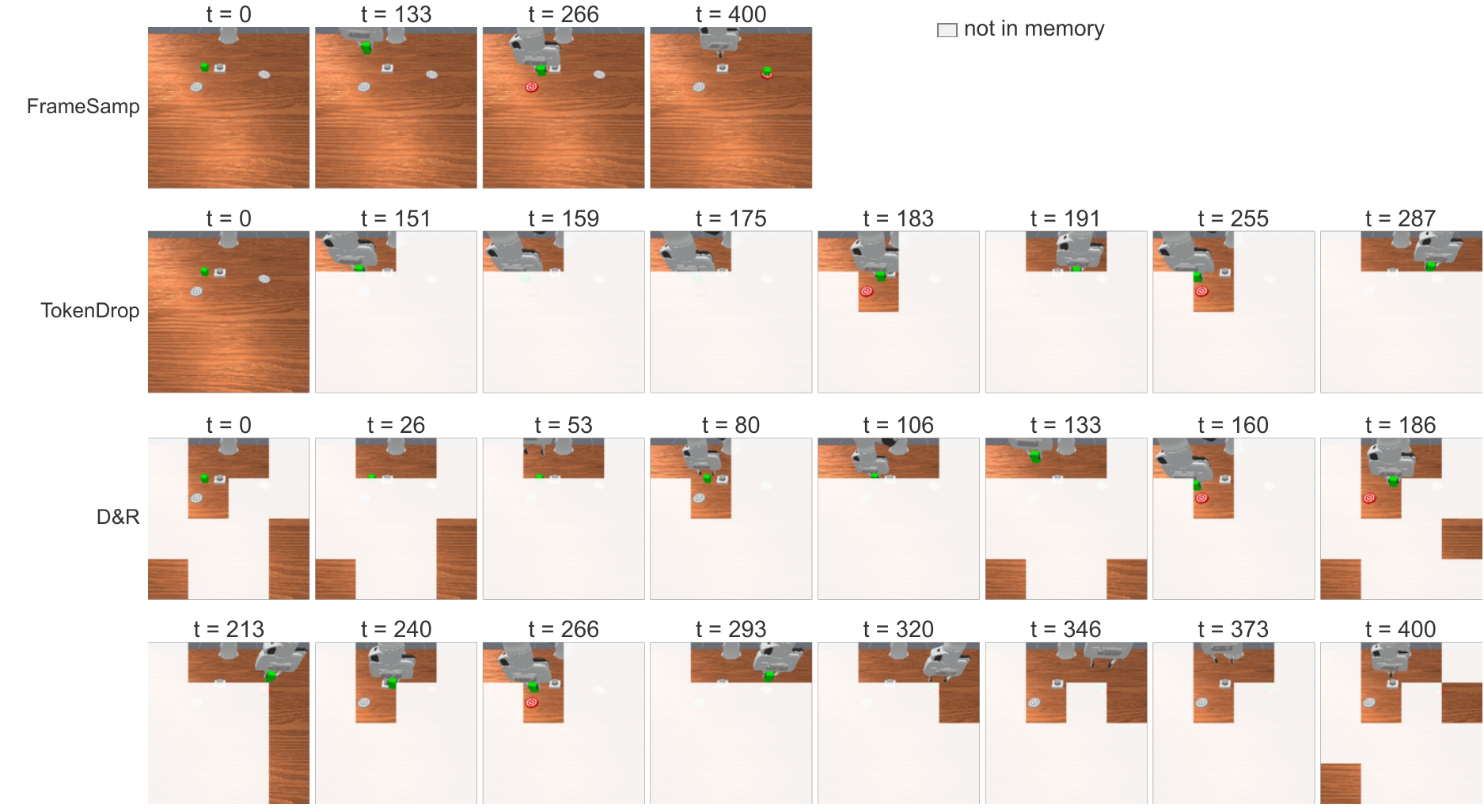}
\vspace{-4pt}
\caption{\textbf{SwingXtimes} (Counting): ``pick up the green cube, move it to the top of the right-side target, then move it to the top of the left-side target, repeating this back-and-forth motion two times, finally press the button to stop''.}
\label{fig:memory_vis_SwingXtimes}
\end{figure}

\begin{figure}[H]
\centering
\vspace{-2pt}
\includegraphics[width=0.78\linewidth]{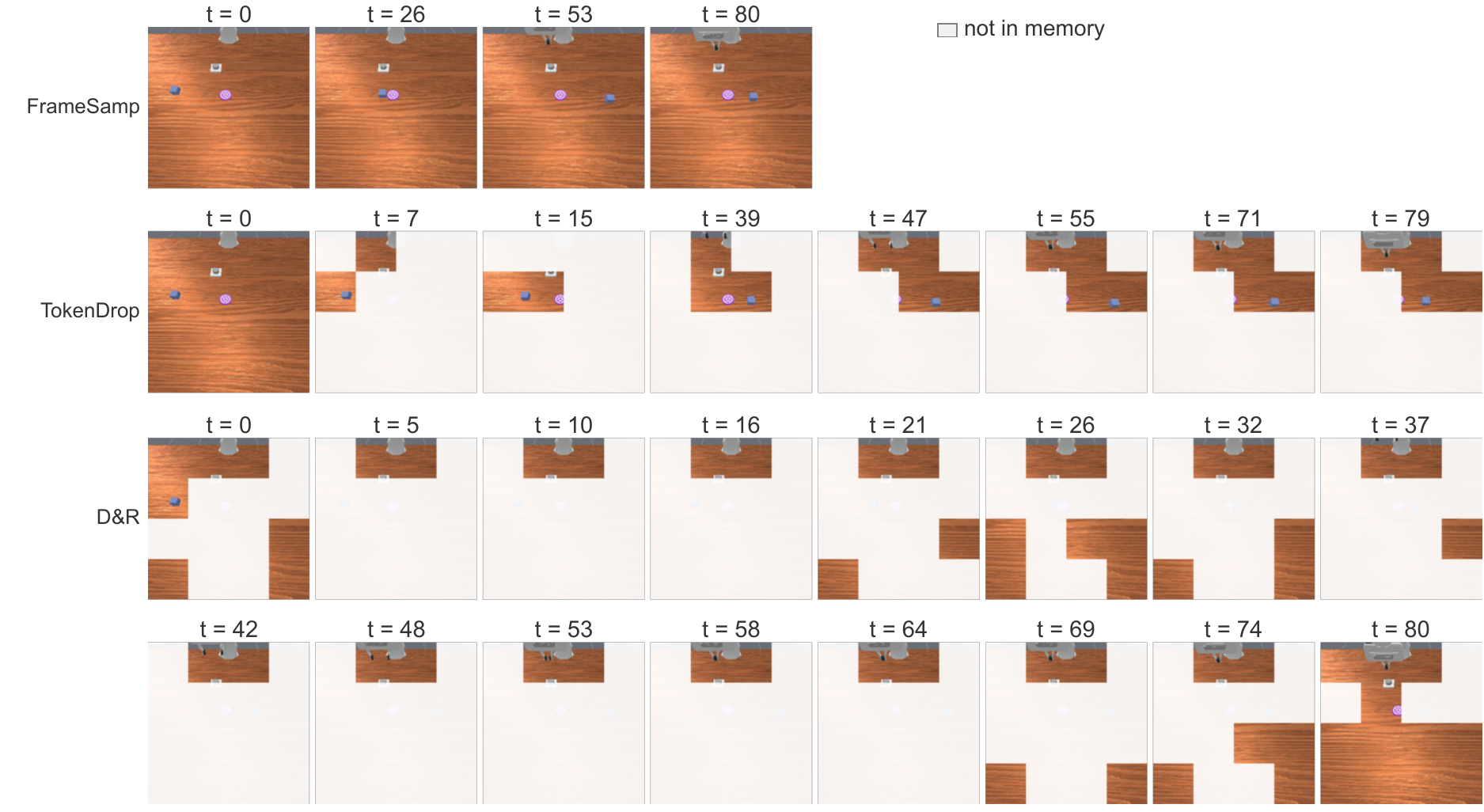}
\vspace{-4pt}
\caption{\textbf{StopCube} (Counting): ``press the button to stop the cube just as it reaches the target for the fourth time''.}
\label{fig:memory_vis_StopCube}
\end{figure}

\begin{figure}[H]
\centering
\vspace{-2pt}
\includegraphics[width=0.78\linewidth]{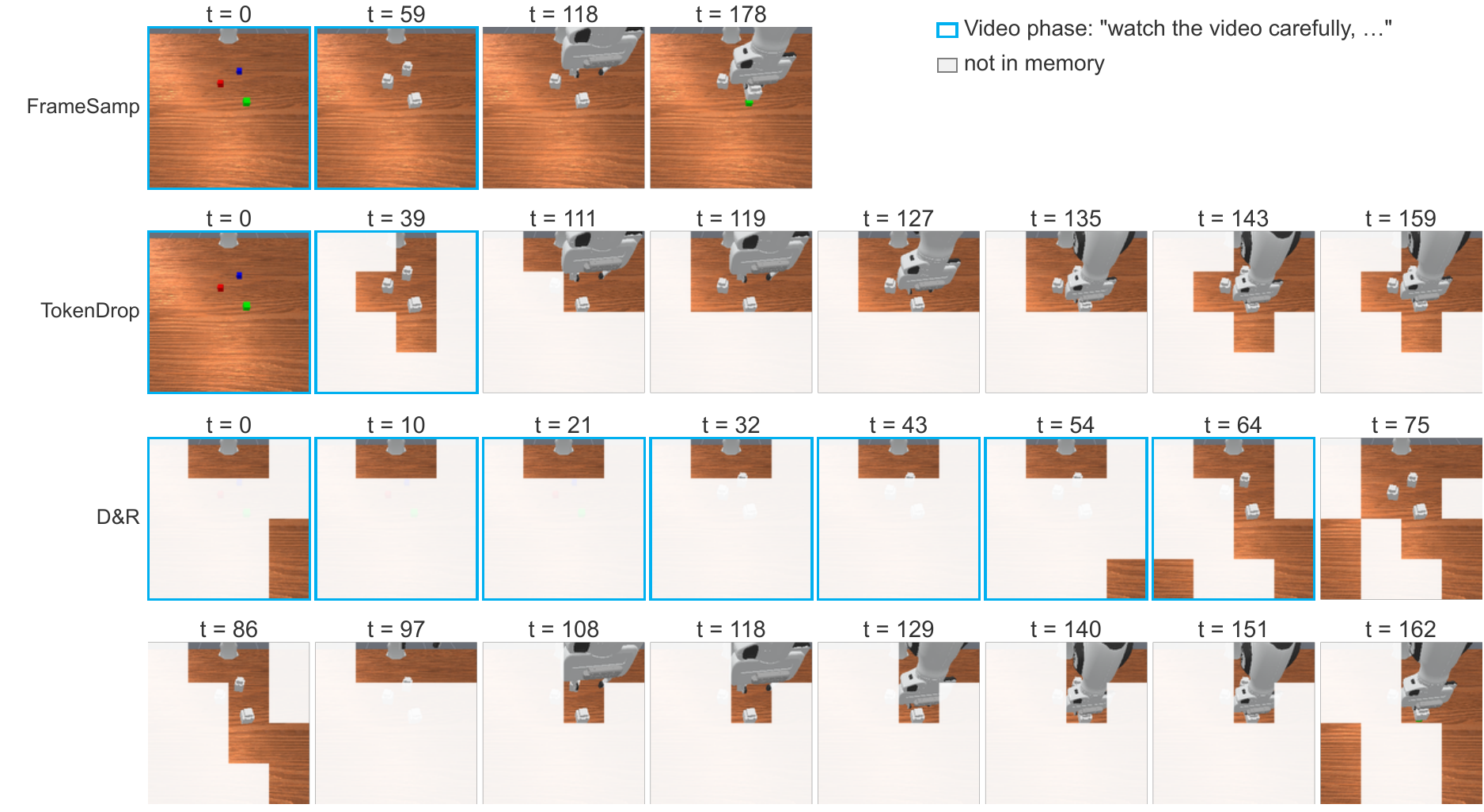}
\vspace{-4pt}
\caption{\textbf{VideoUnmask} (Permanence): ``watch the video carefully, then pick up the container hiding the green cube''.}
\label{fig:memory_vis_VideoUnmask}
\end{figure}

\begin{figure}[H]
\centering
\vspace{-2pt}
\includegraphics[width=0.78\linewidth]{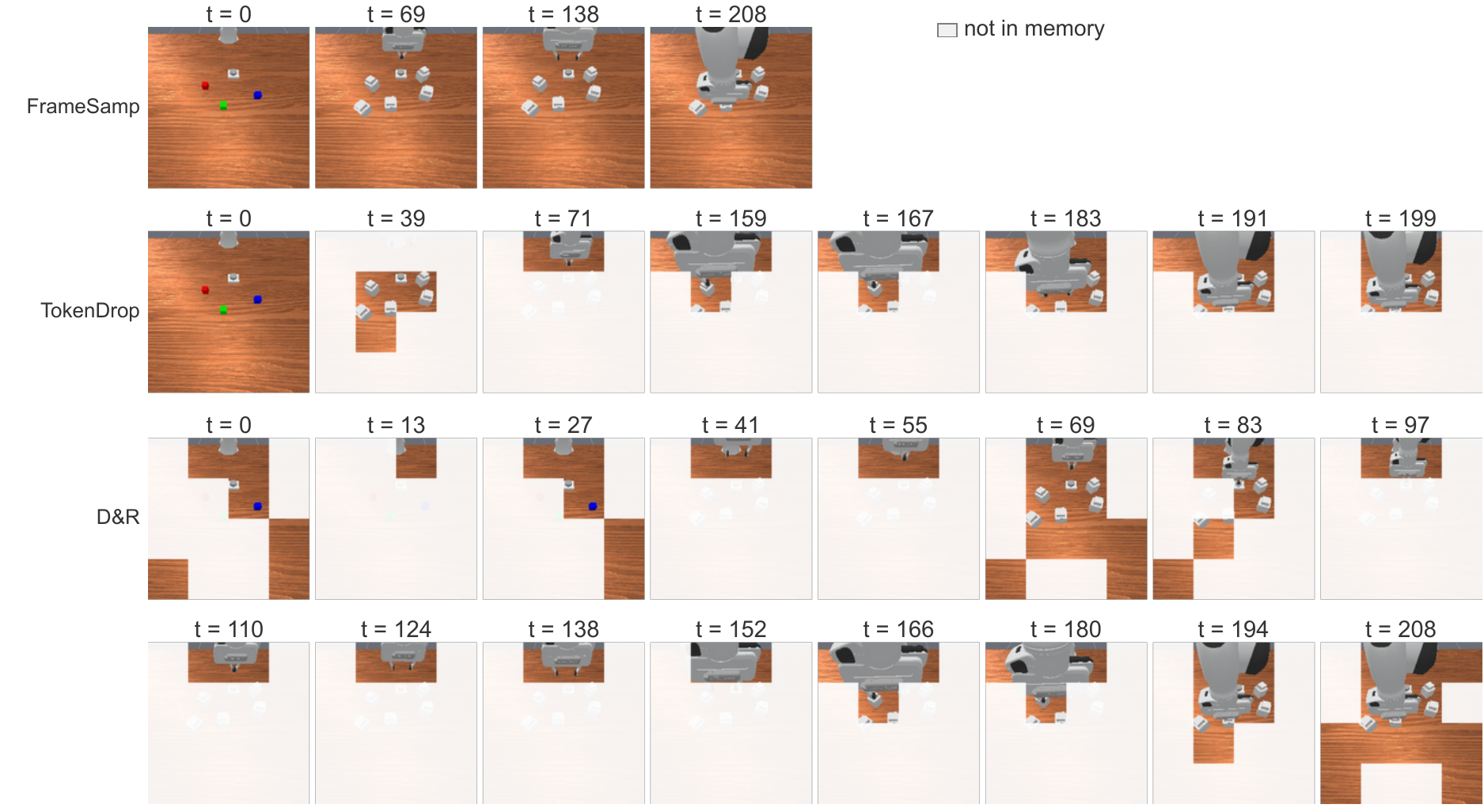}
\vspace{-4pt}
\caption{\textbf{ButtonUnmask} (Permanence): ``first press the button, then pick up the container hiding the green cube''.}
\label{fig:memory_vis_ButtonUnmask}
\end{figure}

\begin{figure}[H]
\centering
\vspace{-2pt}
\includegraphics[width=0.78\linewidth]{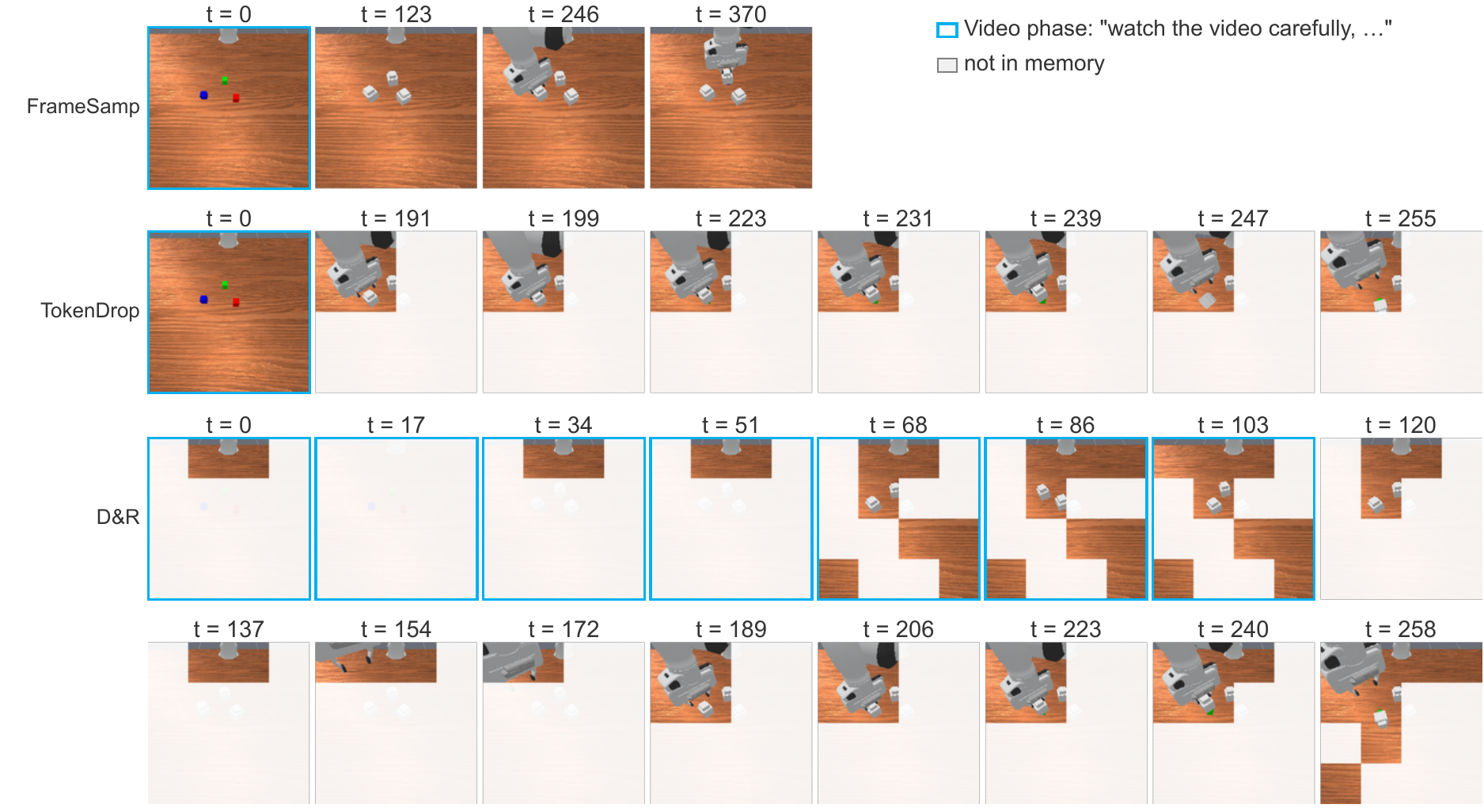}
\vspace{-4pt}
\caption{\textbf{VideoUnmaskSwap} (Permanence): ``watch the video carefully, then pick up the container hiding the green cube''.}
\label{fig:memory_vis_VideoUnmaskSwap}
\end{figure}

\begin{figure}[H]
\centering
\vspace{-2pt}
\includegraphics[width=0.78\linewidth]{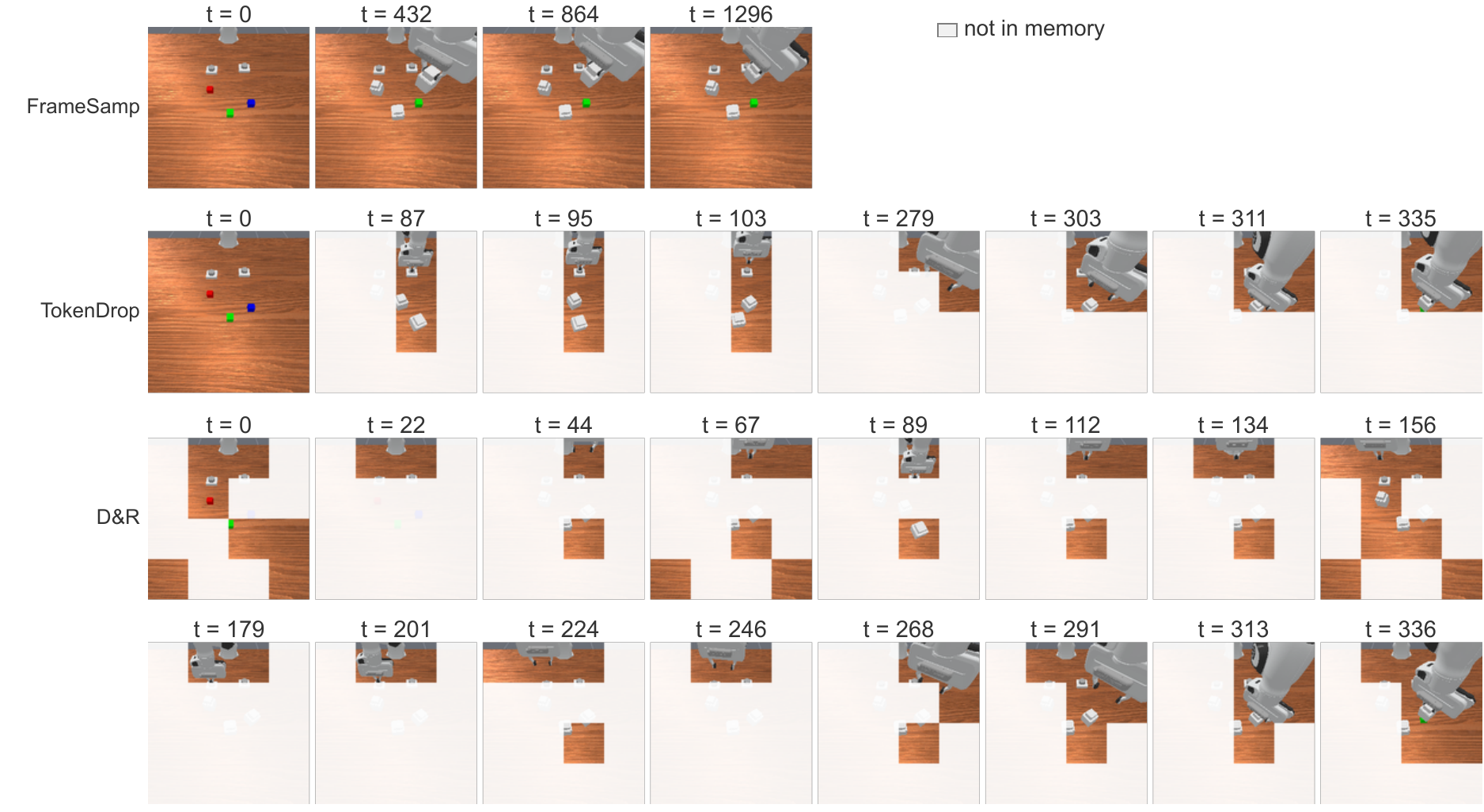}
\vspace{-4pt}
\caption{\textbf{ButtonUnmaskSwap} (Permanence): ``first press both buttons on the table, then pick up the container hiding the green cube''.}
\label{fig:memory_vis_ButtonUnmaskSwap}
\end{figure}

\begin{figure}[H]
\centering
\vspace{-2pt}
\includegraphics[width=0.78\linewidth]{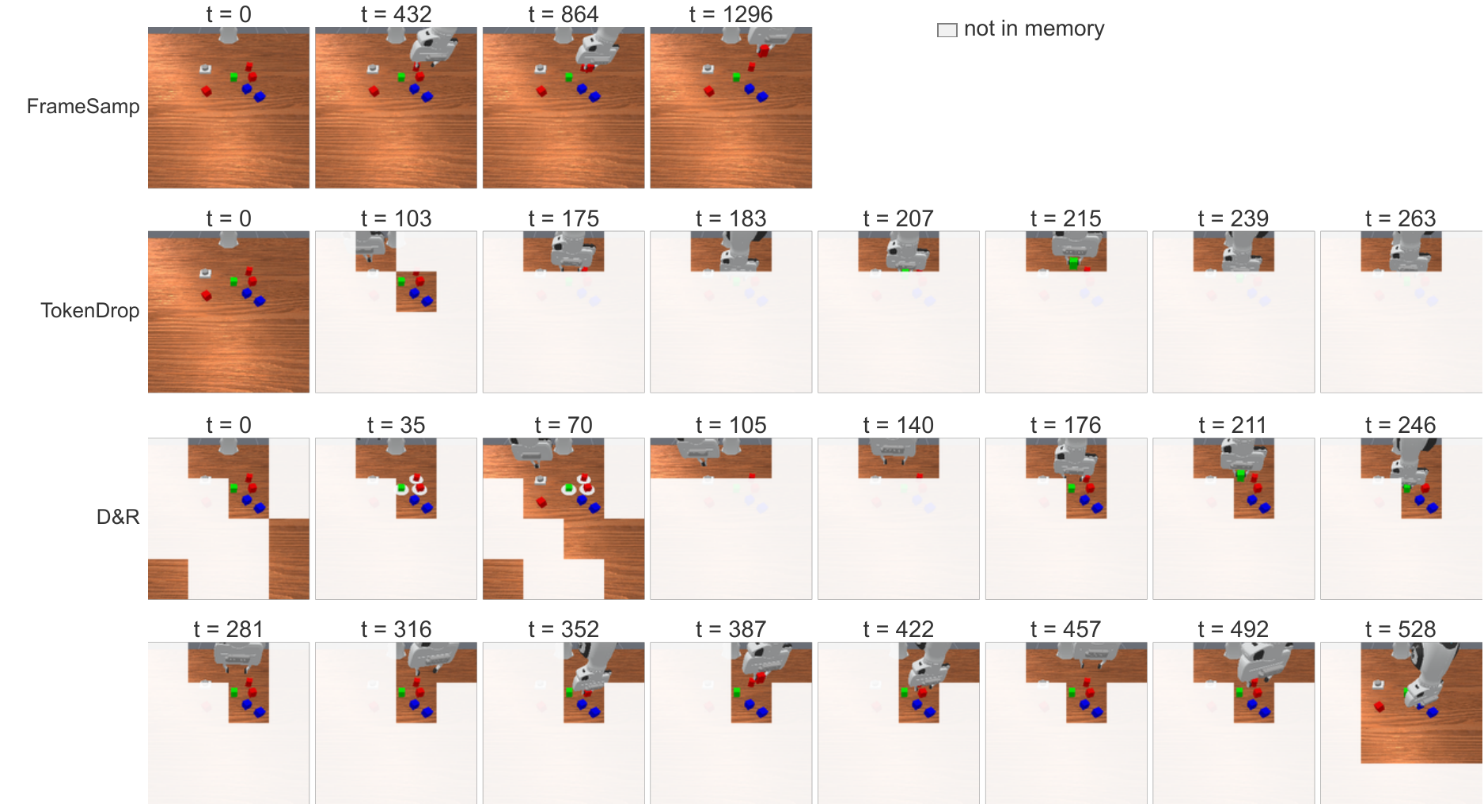}
\vspace{-4pt}
\caption{\textbf{PickHighlight} (Reference): ``first press the button, then pick up all cubes that have been highlighted with white areas on the table''.}
\label{fig:memory_vis_PickHighlight}
\end{figure}

\begin{figure}[H]
\centering
\vspace{-2pt}
\includegraphics[width=0.78\linewidth]{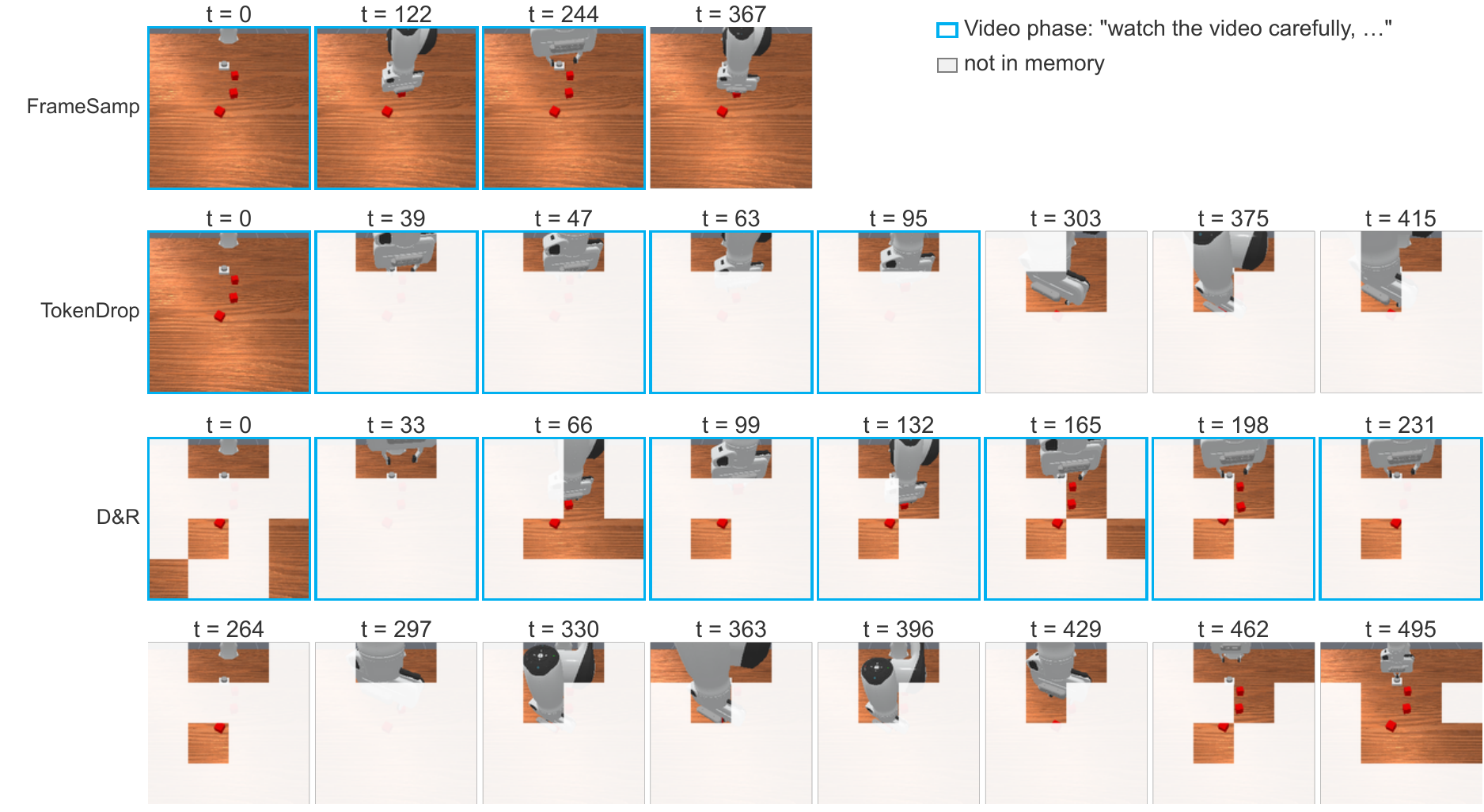}
\vspace{-4pt}
\caption{\textbf{VideoRepick} (Reference): ``watch the video carefully, then pick up the same block that was previously picked up again, finally put it down and press the button to stop''.}
\label{fig:memory_vis_VideoRepick}
\end{figure}

\begin{figure}[H]
\centering
\vspace{-2pt}
\includegraphics[width=0.78\linewidth]{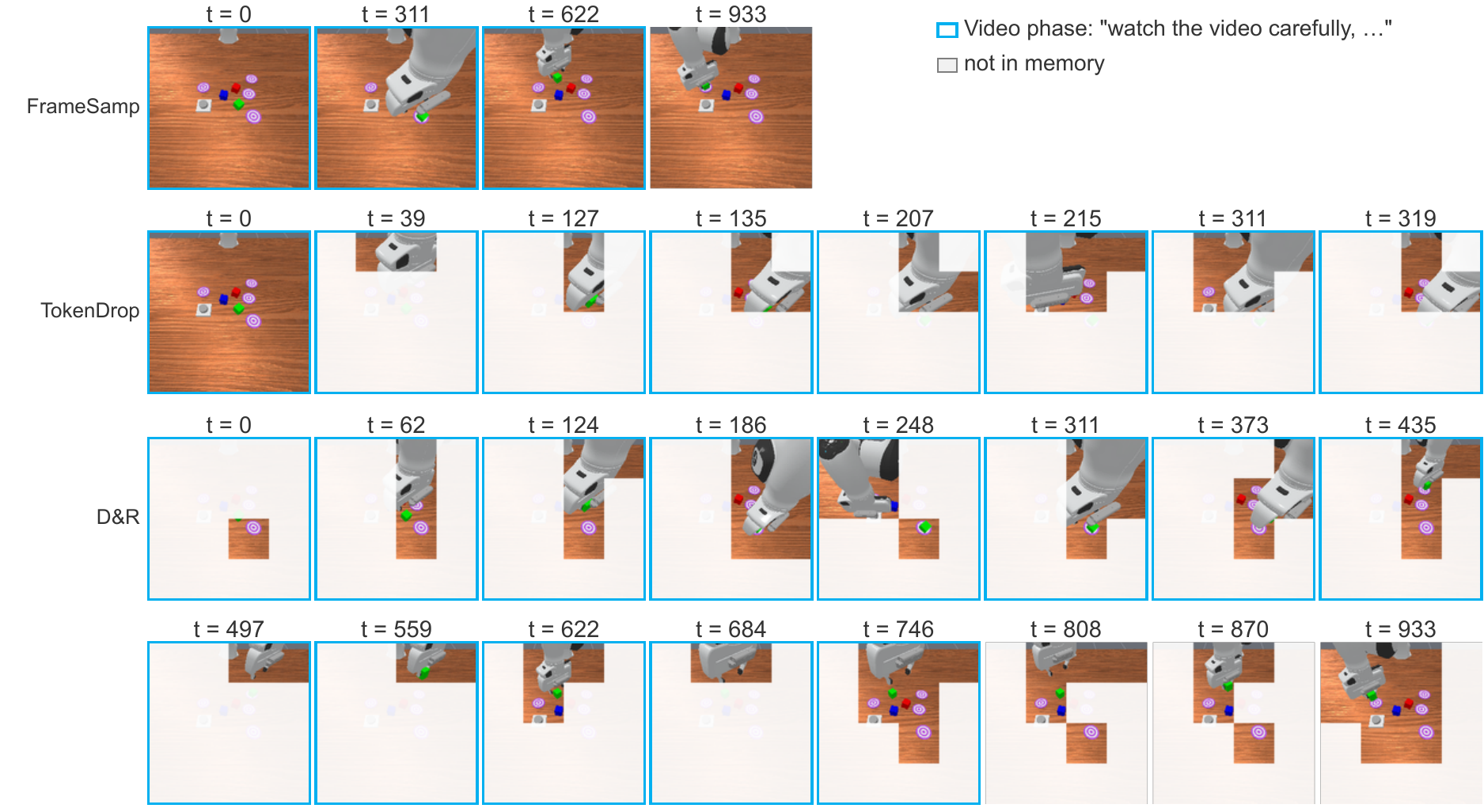}
\vspace{-4pt}
\caption{\textbf{VideoPlaceButton} (Reference): ``watch the video carefully, then place the green cube on the target right before the button was pressed''.}
\label{fig:memory_vis_VideoPlaceButton}
\end{figure}

\begin{figure}[H]
\centering
\vspace{-2pt}
\includegraphics[width=0.78\linewidth]{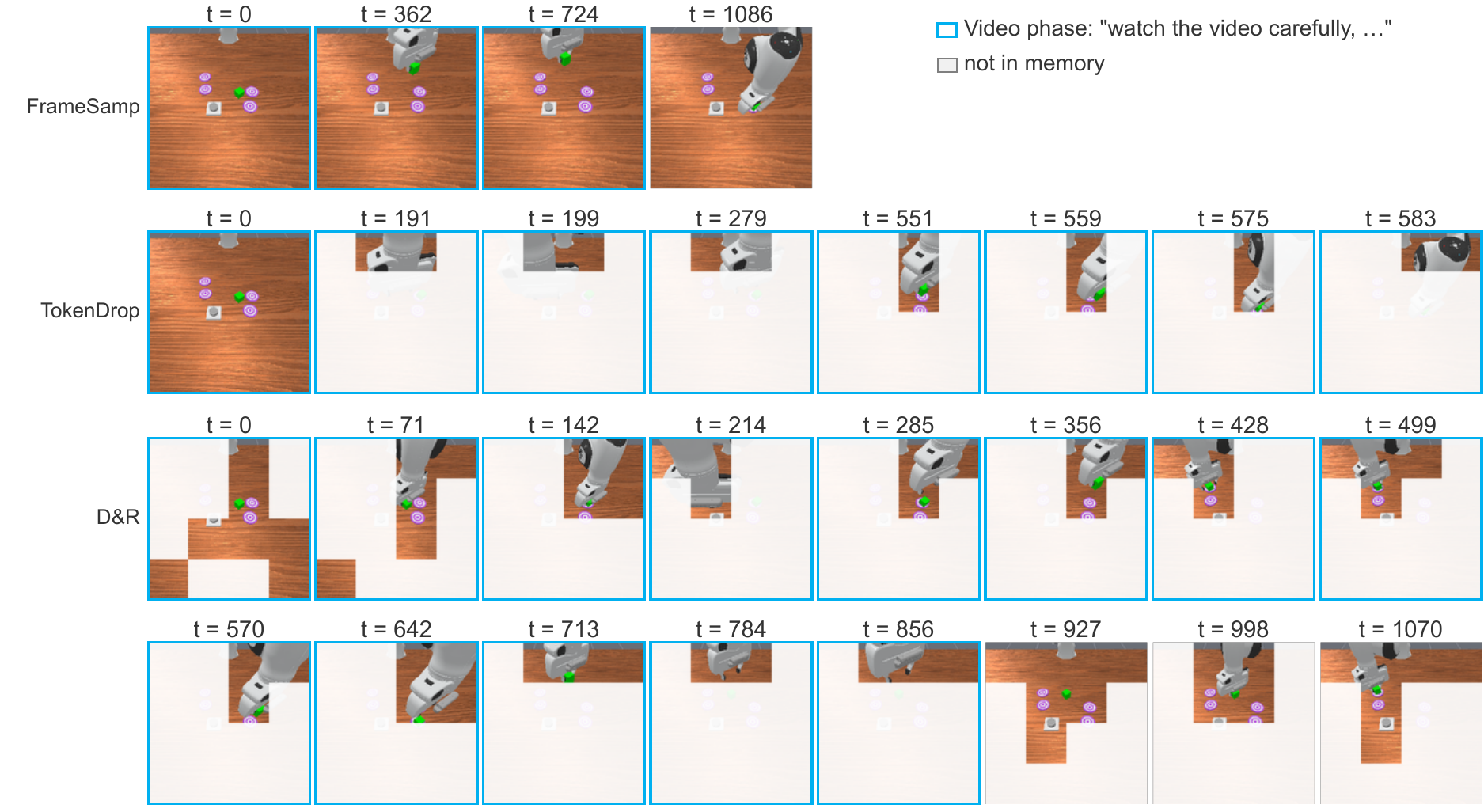}
\vspace{-4pt}
\caption{\textbf{VideoPlaceOrder} (Reference): ``watch the video carefully, then place the green cube on the second target it was previously placed on''.}
\label{fig:memory_vis_VideoPlaceOrder}
\end{figure}

\begin{figure}[H]
\centering
\vspace{-2pt}
\includegraphics[width=0.78\linewidth]{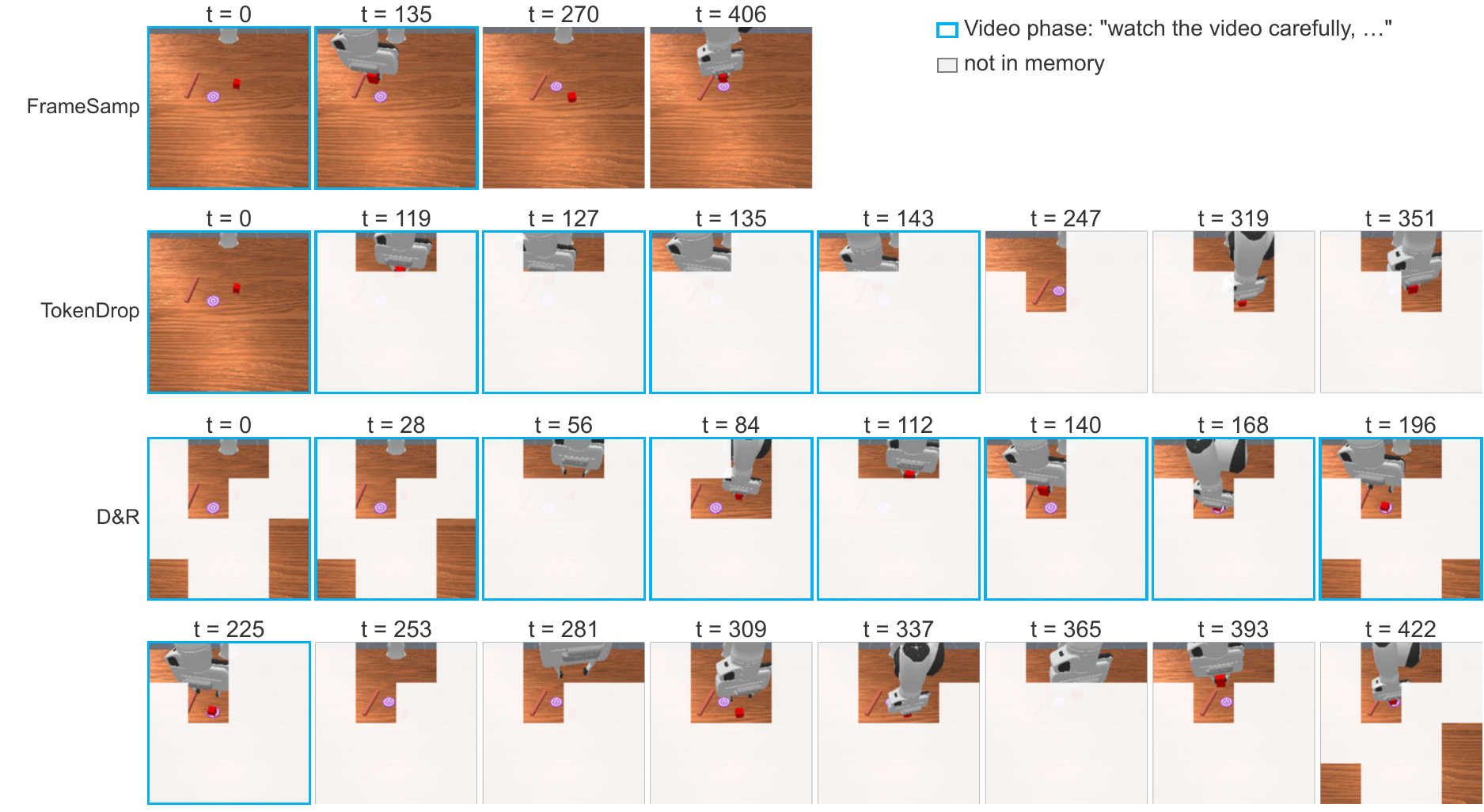}
\vspace{-4pt}
\caption{\textbf{MoveCube} (Imitation): ``watch the video carefully, then move the cube to the target in the same manner as before''.}
\label{fig:memory_vis_MoveCube}
\end{figure}

\begin{figure}[H]
\centering
\vspace{-2pt}
\includegraphics[width=0.78\linewidth]{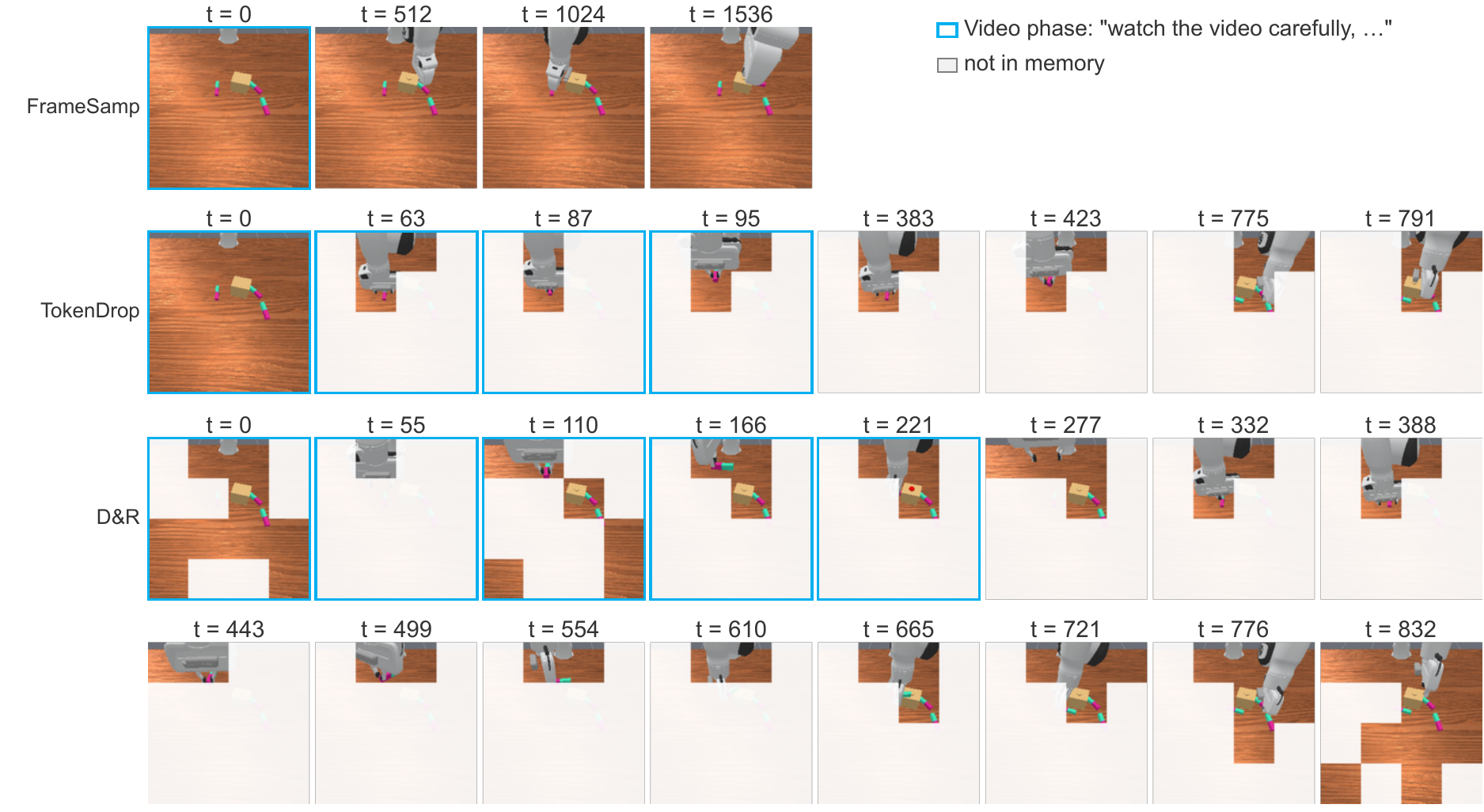}
\vspace{-4pt}
\caption{\textbf{InsertPeg} (Imitation): ``watch the video carefully, then grasp the same end of the same peg you've picked before and insert it into the same side of the box''.}
\label{fig:memory_vis_InsertPeg}
\end{figure}

\begin{figure}[H]
\centering
\vspace{-2pt}
\includegraphics[width=0.78\linewidth]{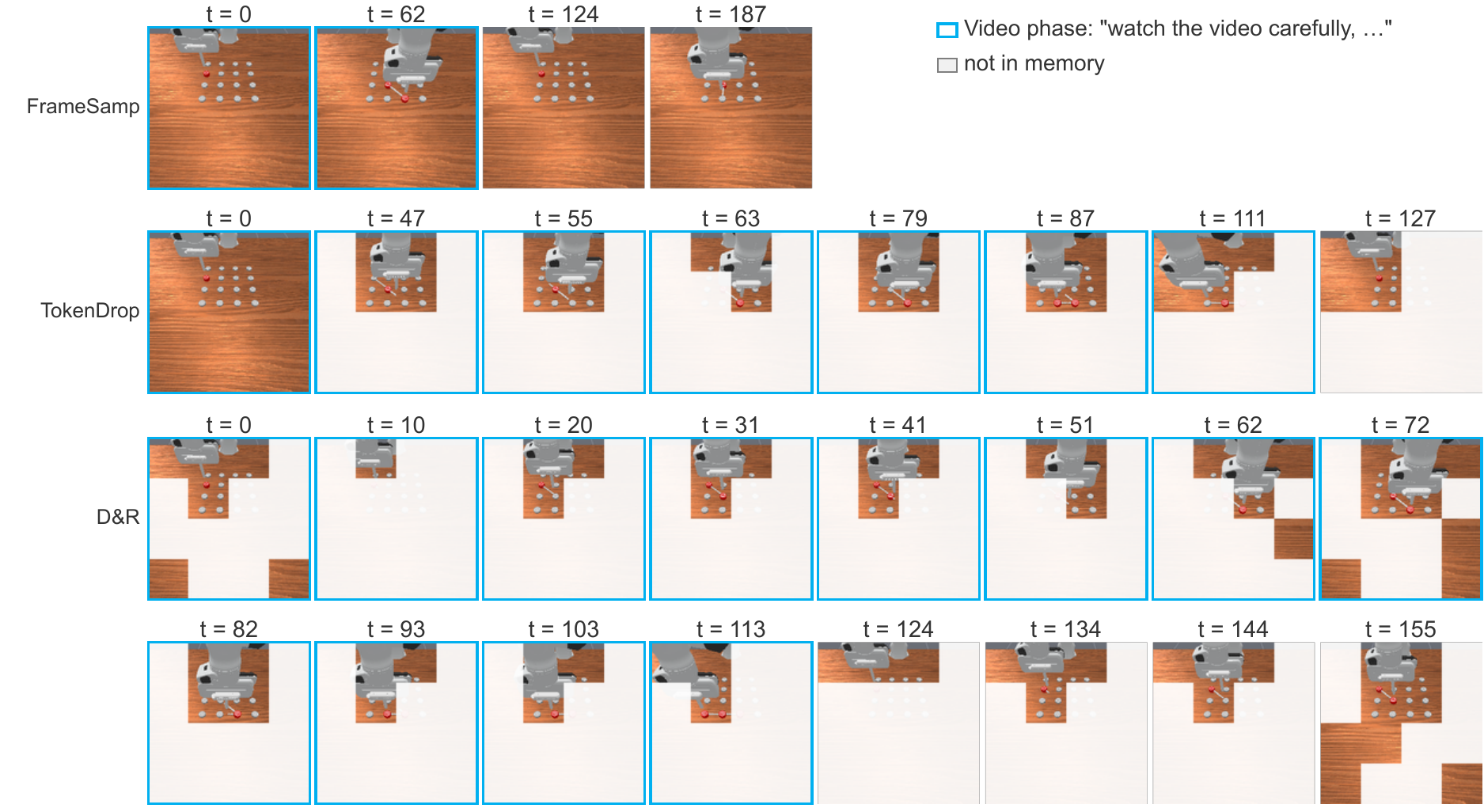}
\vspace{-4pt}
\caption{\textbf{PatternLock} (Imitation): ``watch the video carefully, then use the stick attached to the robot to retrace the same pattern''.}
\label{fig:memory_vis_PatternLock}
\end{figure}

\begin{figure}[H]
\centering
\vspace{-2pt}
\includegraphics[width=0.78\linewidth]{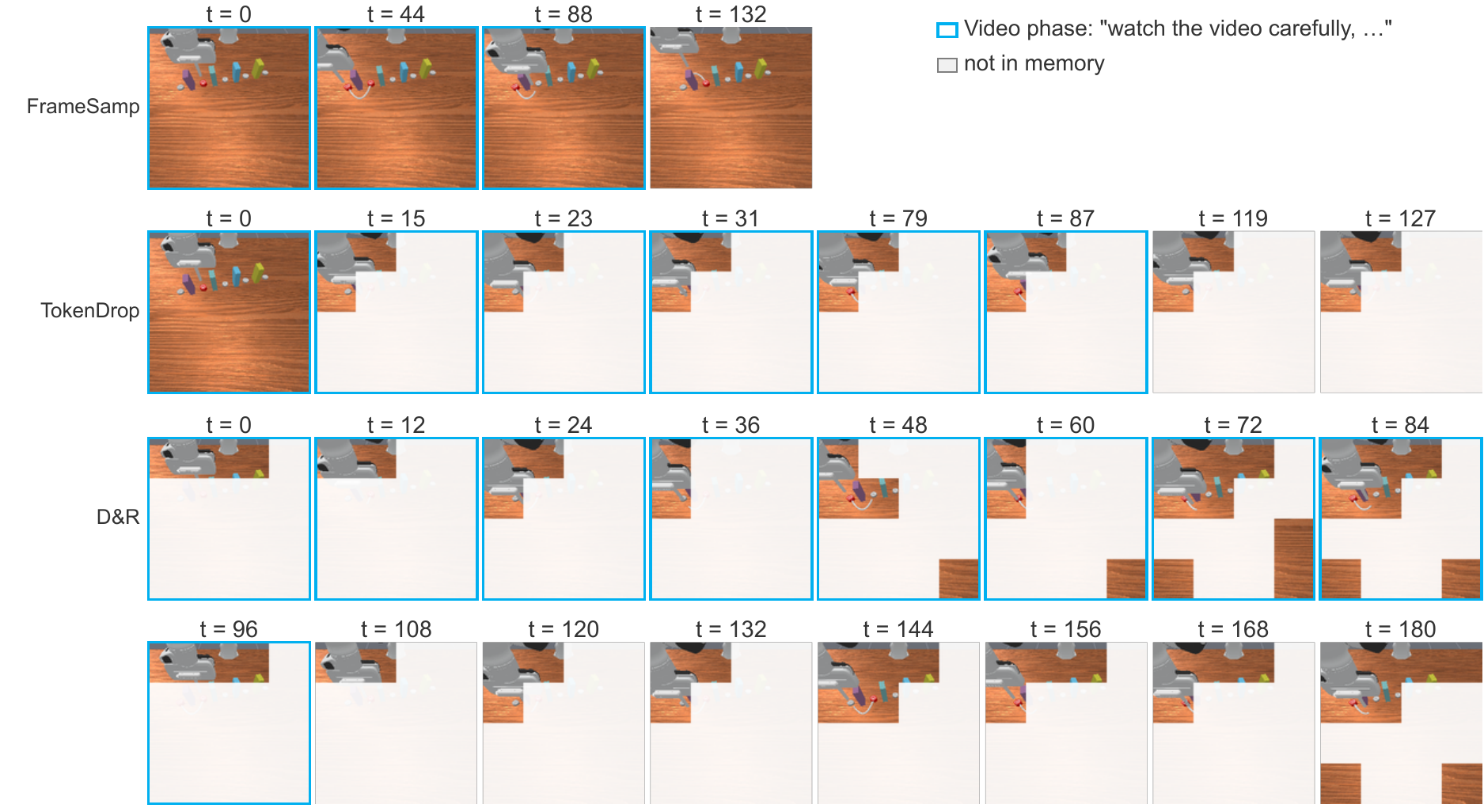}
\vspace{-4pt}
\caption{\textbf{RouteStick} (Imitation): ``watch the video carefully, then use the stick attached to the robot to navigate around the sticks on the table, following the same path''.}
\label{fig:memory_vis_RouteStick}
\end{figure}

Figure~\ref{fig:memory_vis} in the main text visualises the memory of one MoveCube episode. Figures~\ref{fig:memory_vis_BinFill}--\ref{fig:memory_vis_RouteStick} repeat the visualisation for every RoboMME task.

\clearpage
\subsection{Real robot experiment}\label{app:real-robot}
\begin{figure}[h]
\centering
\includegraphics[width=\linewidth]{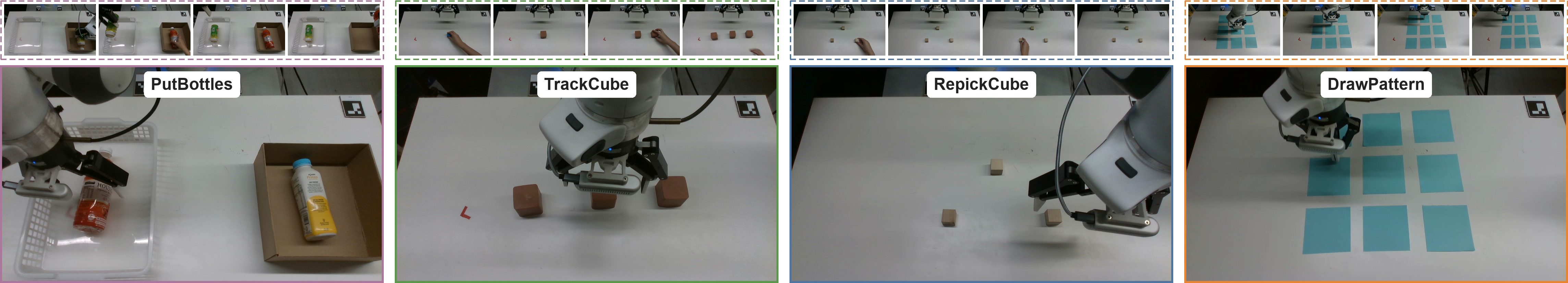}
\caption{The four real-world tasks, recorded on our Franka Panda: PutBottles (Counting), TrackCube (Permanence), RepickCube (Reference) and DrawPattern (Imitation). For each task, the upper dashed box shows frames from the earlier history (the video the policy watched; for PutBottles, earlier frames of the episode, including a human intervention), and the lower box shows the front-camera observation during execution.}
\label{fig:real-robot-tasks}
\end{figure}

Following~\citet{dai2026robomme}, we evaluate \ours on a real robot with four tasks, PutBottles, TrackCube, RepickCube, and DrawPattern, which mirror the simulation tasks BinFill, VideoUnmask, VideoRepick, and PatternLock, respectively (Figure~\ref{fig:real-robot-tasks}).
\begin{itemize}[leftmargin=*,nosep]
\item PutBottles (Counting). The robot transfers $N\in\{1,2\}$ bottles from a basket to a bin. During execution, a human may intervene by removing placed bottles or adding new ones, so the robot cannot rely on the current observation alone and must track its progress over time.
\item TrackCube (Permanence). Given a pre-recorded video in which a human hides cubes under identical cups, the robot must lift the cup that hides the cube of the named colour; $N\in\{3,4\}$ is the number of cups (one cube under one of three cups, or three cubes under three of four cups).
\item RepickCube (Reference). Given a pre-recorded video in which a human picks up two of the $N\in\{3,4\}$ cubes on the table, the robot must pick up the same two cubes.
\item DrawPattern (Imitation). Given a pre-recorded video in which a human traces a path of $N\in\{2,3,4\}$ moves on a $3\times3$ grid, the robot must reproduce the same path.
\end{itemize}

\begin{figure}[h]
\centering
\begin{tikzpicture}[font=\small, line cap=round]
  \node[anchor=south west, inner sep=0] (img) {\includegraphics[width=0.4\linewidth]{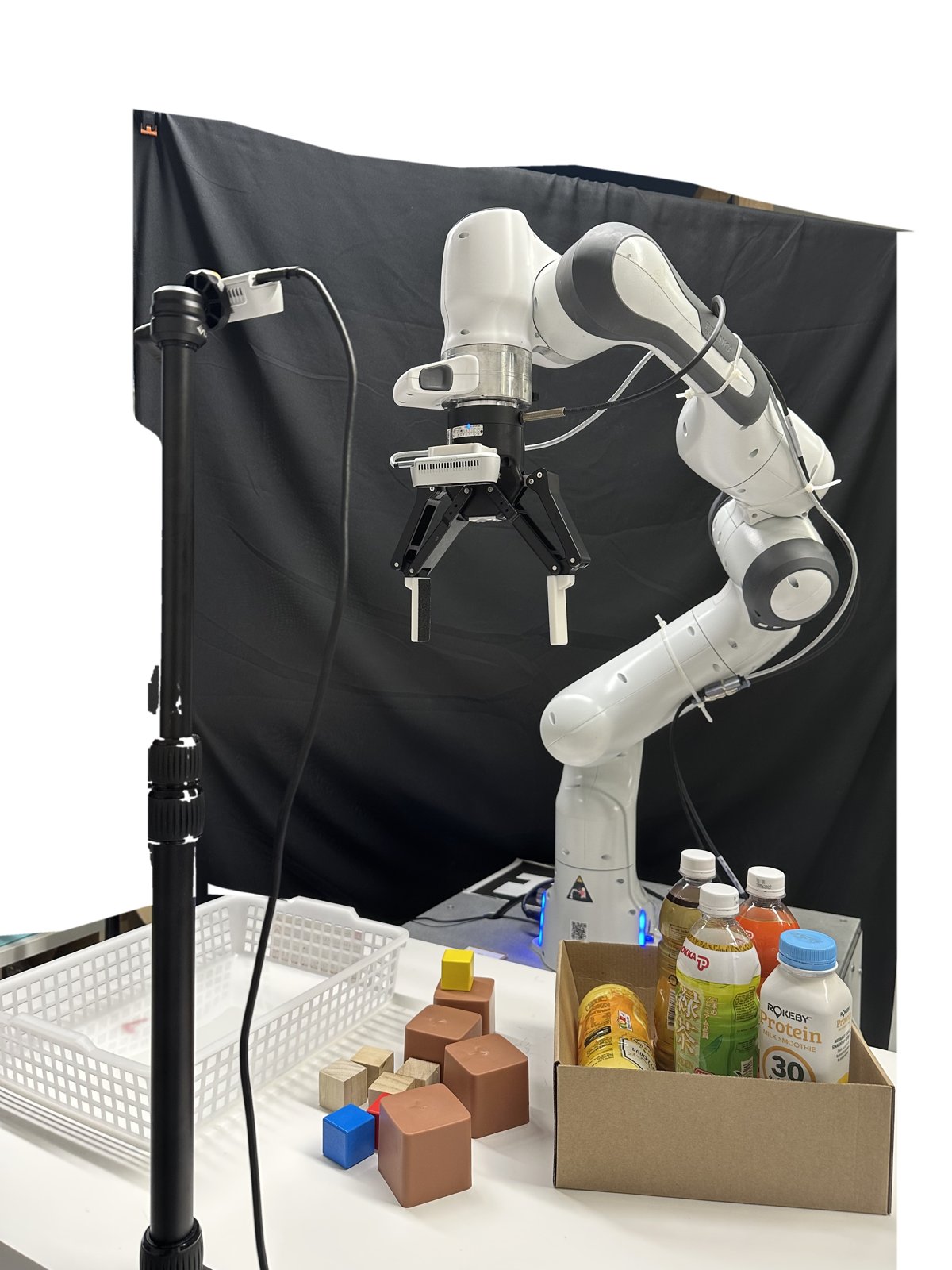}};
  \begin{scope}[x={(img.south east)}, y={(img.north west)},
                box/.style={draw=red, line width=1.2pt, rounded corners=3pt},
                arr/.style={-{Stealth[length=6pt, width=5pt]}, draw=red, line width=0.9pt, dash pattern=on 4pt off 2.5pt},
                lbl/.style={text=black, align=center, inner sep=1pt}]
    \draw[box] (0.20, 0.735) rectangle (0.32, 0.805);
    \node[lbl, anchor=east] (front) at (-0.12, 0.70) {Front\\Camera};
    \draw[arr] (front.east) -- (0.20, 0.755);
    \draw[box] (0.41, 0.595) rectangle (0.545, 0.67);
    \node[lbl, anchor=west] (wrist) at (1.12, 0.56) {Wrist\\Camera};
    \draw[arr] (wrist.west) -- (0.545, 0.615);
  \end{scope}
\end{tikzpicture}
\caption{Real-robot platform. A 7-DoF Franka Emika Panda with a Robotiq 2F-140 gripper and two RealSense cameras: a wrist camera mounted on the gripper and a front camera on a tripod facing the workspace. The objects of the four tasks (baskets, cubes, cups, and bottles) are laid out on the table.}
\label{fig:real-robot-setup}
\end{figure}

\paragraph{Protocol.} We use a 7-DoF Franka Emika Panda arm with a Robotiq 2F-140 gripper, controlled through deoxys~\citep{zhu2022viola} in joint-position mode at $1000$\,Hz, and two Intel RealSense cameras: a D435i mounted on the wrist and a D435 on a tripod in front of the workspace (Figure~\ref{fig:real-robot-setup}). The memory is built from the front view only; the wrist view is fed to the policy as a current observation. Demonstrations are collected by teleoperation at $15$\,Hz, $100$ per task, and, as in simulation, all methods are trained in a multi-task learning setting on the four tasks. Each task is evaluated over $10$ rollouts per method; we compare $\pi_{0.5}$, FrameSamp with a budget of $K=2048$ tokens, and \ours
with $K=1024$ tokens drawn from a pool of $|\mathcal H|=2048$. The budgets are
larger than in simulation because the real-robot episodes are longer. All other
hyperparameters are unchanged, except that training runs for $50$k steps. Table~\ref{tab:realworld} reports the results.

\paragraph{Results.} Table~\ref{tab:realworld} reports the successes over the $10$ trials per task. We inspected every failed trial using the rollout videos and the robot logs (gripper width, grasp poses and stop reasons).

\begin{description}[leftmargin=1.2em, labelsep=0.4em, itemsep=3pt, topsep=3pt, parsep=0pt]
\item[(1) PutBottles.] Without memory, $\pi_{0.5}$ cannot count. It overshoots in four trials (a further grasp after the last bottle), stops one bottle short in two, and never transfers a bottle in three. FrameSamp remembers the wrong count in three trials, stops short in one and never transfers a bottle in one. In one trial each, $\pi_{0.5}$ and FrameSamp place both bottles for $N{=}2$ but the trial is still judged a failure. \ours fails once, grasping a third bottle after placing both for $N{=}2$.
\item[(2) TrackCube.] All eight failures of $\pi_{0.5}$ uncover a wrong cup. Both FrameSamp failures do the same, because the memory holds the wrong frame. \ours lifts the correct cup in all ten trials; one is logged as a failure because the run was stopped before the automatic put-back.
\item[(3) RepickCube.] $\pi_{0.5}$ never picks the right cubes, and two trials end in a safety stop. Three of FrameSamp's four failures pick the first cube correctly and then fail from the wrong memory, heading for a wrong cube or stopping; the fourth grasps no cube. The only \ours failure picks the first cube and then a wrong second one.
\item[(4) DrawPattern.] $\pi_{0.5}$ makes the same leftward move in eight of its nine failed trials, whatever the demonstration, and four of them end with a safety stop. FrameSamp reproduces the first two moves, but as the error in its memory accumulates it goes wrong at the third in four trials; in the other two it adds a move at the end or ends imprecisely. The two \ours failures come from a wrong movement direction, at the third move in one trial and at the first in the other.
\end{description}

\end{document}